\documentclass{article}

\PassOptionsToPackage{authoryear,round}{natbib}
\usepackage[preprint]{neurips_2025}

\usepackage[utf8]{inputenc} % allow utf-8 input
\usepackage[T1]{fontenc}    % use 8-bit T1 fonts
\usepackage{hyperref}       % hyperlinks
\usepackage{url}            % simple URL typesetting
\usepackage{booktabs}       % professional-quality tables
\usepackage{amsfonts}       % blackboard math symbols
\usepackage{nicefrac}       % compact symbols for 1/2, etc.
\usepackage{microtype}      % microtypography
\usepackage{xcolor}         % colors

\usepackage{booktabs}
\usepackage{xspace}
\usepackage{colortbl}
\usepackage{multirow}
\usepackage{microtype}
\usepackage{amsmath}
\usepackage{mathtools}
\usepackage{amsthm}
\usepackage{array}
\usepackage{calc}
\usepackage{graphicx}
\usepackage{hyperref}
\usepackage{footnote}
\usepackage{enumitem}

\usepackage{newfloat}
\usepackage{listings}
\usepackage{subcaption}  
\usepackage{xcolor}
\usepackage{seqsplit}
\usepackage{url}

\newcommand{\model}{ProteinGPT\xspace}
\newcommand{\dataset}{ProteinQA\xspace}
\newcommand{\datasize}{132,092\xspace}

\usepackage{ifthen} 
\newboolean{forKDD}

\title{ProteinGPT: Multimodal LLM for Protein Property Prediction and Structure Understanding}

\author{Yijia Xiao$^{1}$, Edward Sun$^{1}$, Yiqiao Jin$^{2}$, Qifan Wang$^{3}$, Wei Wang$^{1}$ \\
$^{1}$University of California, Los Angeles, 
$^{2}$Georgia Institute of Technology, 
$^{3}$Meta AI \\
\texttt{\{yijia.xiao,wei.wang\}@cs.ucla.edu}, 
\texttt{edwardsun12895@g.ucla.edu}, \\
\texttt{yjin328@gatech.edu}, \texttt{wqfcr@fb.com}
}

\begin{document}

\maketitle

\begin{abstract}
Understanding biological processes, drug development, and biotechnological advancements requires a detailed analysis of protein structures and functions, a task that is inherently complex and time-consuming in traditional protein research. 
% Protein research is key to . However, manual analysis of protein structures and sequences is complex and time-consuming. 
To streamline this process, we introduce \model, a state-of-the-art multimodal large language model for proteins, which allows users to upload protein sequences and/or structures for comprehensive proteins analysis and responsive inquiries.  
ProteinGPT seamlessly integrates protein sequence and structure encoders with linear projection layers to ensure precise representation adaptation. It leverages a large language model (LLM) to generate accurate and contextually relevant responses. 
To train \model, we construct a large-scale dataset of \datasize proteins, each annotated with 20-30 property tags and 5-10
QA pairs per protein, and optimized the instruction-tuning process using GPT-4o. % This innovative system ensures accurate alignment between the user uploaded data and prompts, simplifying protein analysis. 
Experiments demonstrate that \model effectively generates informative responses to protein-related questions, achieving high performance on both semantic and lexical metrics. It significantly outperforms baseline models and general-purpose LLMs in understanding and responding to protein-related queries. 
Our code and data are available at \url{https://github.com/ProteinGPT/ProteinGPT}. 
\end{abstract}

\section{Introduction}

Proteins are fundamental molecular building blocks of life, playing critical roles in biological processes~\citep{kitadai2018origins,xiao2025protein}. 
Understanding their structure, functions, and interactions is vital for advancements in drug discovery~\citep{Teague2003}, healthcare~\citep{world2007protein}, and biological/medical engineering~\citep{Kobsa_Saltzman}. 
Recent breakthroughs in machine-learning-based protein structure and function prediction~\citep{ESM-2} have significantly accelerated biological research by reducing the reliance on traditional labor-intensive laboratory experiments and literature search. 

\noindent \textbf{Challenges.} As proteins can be represented by strings of characters, each corresponding to an amino acid from an alphabet of 20 letters, recent advancements in Large Language Models (LLMs) have naturally extended to protein research. 
% Recent success of Large Language Models (LLMs) have naturally facilitated the application of these NLP advancements to protein research, as proteins are represented by a string of characters with each character representing an amino acid from an alphabet of 20 characters. 
Existing protein LLMs such as ProtST~\citep{xu2023protst}, ProteinChat~\citep{guo2023proteinchat}, and ProtChatGPT~\citep{wang2024protchatgpt} % align protein sequence or structure data with textual descriptions to create LLMs for proteins. 
% have spearheaded ways to align protein structure or sequence data along with text abstract descriptions of each protein to LLMs. 
focus primarily on sequence-based or structure-based modeling, limiting their ability to generate holistic protein insights from multiple modalities. For instance, protein sequences can reveal evolutionary information, functional sites, and sequence-structure relationships, while protein structures provide critical insights into spatial arrangement, structural dynamics, binding sites, and stability. 
% \yq{Add this and need proofread:
Applying multimodal LLMs to protein modeling is non-trivial due to the challenge in aligning diverse modalities, such as textual descriptions, protein sequences, and protein structures. 
% Although LLMs excel in language generation and zero-shot transfer..
Meanwhile, direct end-to-end retraining for protein LLMs is usually impractical due to extensive requirements for annotated data.
% the key challenge of transferring LLMs to protein research lies in the modality gap between protein structures and biomedical texts. Despite the strong language generation and zero-shot transfer abilities of LLMs, directly retraining them end-to-end for protein spe- cialization appears to be impractical due to the massive num- ber of parameters and data requirements.

% ProtST and ProteinChat rely solely on one modality (protein sequence or structure). This approach may overlook essential insights in the other modality. While ProtChatGPT incorporates both, it is aligned via a PLP-former on a full abstract of the protein, which introduces extra complexity and could cause unnecessarily verbose responses. 

\begin{figure*}[ht]
% \vskip 0.2in
\begin{center}
\centerline{\includegraphics[width=\linewidth]{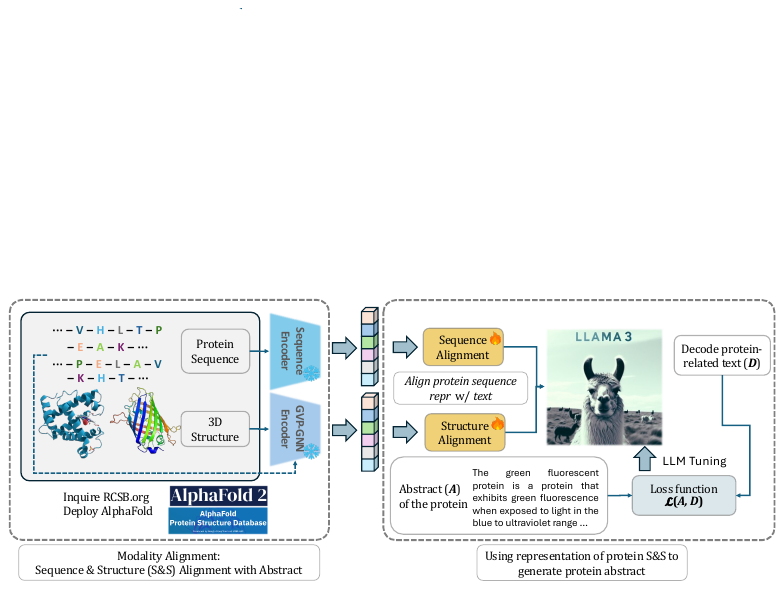}}
\vspace{-2mm}
\caption{\raisebox{-1ex}{\includegraphics[height=24pt]{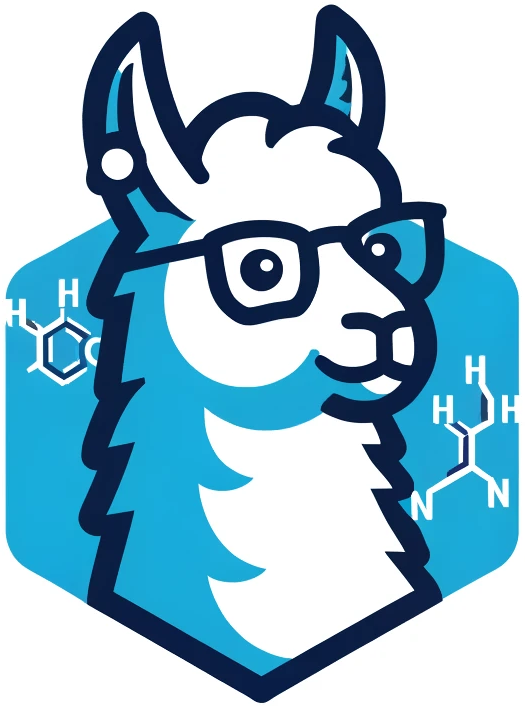}}\hspace{-0.1em} \model Modality Fusion \& Alignment Stage: we freeze the encoder blocks and train the linear project layer to learn how to align protein structure and protein sequence representations with text. In the alignment stage, the input to the training is only the projected protein representation. No text prompts are incorporated in this stage.}
\label{fig:schema_ma}
\end{center}
\vspace{-8mm}
\end{figure*}

\noindent \textbf{Our Work.} We propose \model, a protein LLM that allows researchers to upload protein sequences and/or structures (via fasta or PDB files) and ask natural language questions. \model consists of four major components: a protein sequence encoder, a protein structure encoder, a projection layer, and an LLM. 

The \emph{protein sequence encoder} is based on the ESM-2 (Evolutionary Scale Modeling 2)~\citep{ESM-2} model variant \verb|esm2_t36_3B_UR50D| with 36 transformer layers and 3 billion parameters. Pretrained on \verb|UniRef50/D|~\citep{suzek2015uniref}, a comprehensive protein database that clusters sequences with at least 50\% sequence identity and 80\% coverage, 
% strategically samples \verb|UniRef100| sequences from \verb|UniRef50| clusters to 
this model ensures sequence diversity and informativeness in encoding. 
% data samples \verb|UniRef100| sequences evenly across \verb|UniRef50| clusters to enhance sequence diversity, making the encoding informative. 
The \emph{protein structure encoder}, \verb|esm_if1_gvp4_t16_142M_UR50|, is an inverse folding model that incorporates a geometric input processing layer paired with a seq2seq transformer~\citep{bahdanau2014neural}. Trained on 12 million structures predicted by AlphaFold2~\citep{JumperAlphaFold2021}, the model effectively captures protein structural information. 
% The protein structure encoder, \verb|esm_if1_gvp4_t16_142M_UR50|, captures the structural information of the protein. The structure encoder is an inverse folding model that features a geometric input processing layer with a seq2seq transformer. Trained on 12 million structures predicted by AlphaFold2 \citep{JumperAlphaFold2021}, the model can effectively capture the structural information of the protein. 
% To encode the proteins, we utilize the ESM-2 (Evolutionary Scale Modeling 2) protein language models. For protein sequence encoding, we use \verb|esm2_t36_3B_UR50D|, which features 36 transformer layers and 3 billion parameters, trained on the Protein \verb|UniRef50/D| database to enhance sequence diversity. 
% which samples \verb|UniRef100| sequences evenly across \verb|UniRef50| clusters to enhance sequence diversity, making the encoding informative. 
% features a geometric input processing layer with a seq2seq transformer trained on 12 million structures predicted by AlphaFold2 \citep{JumperAlphaFold2021}. 

To bridge these encoders with the LLM, we introduce a \emph{projection layer} that aligns their embeddings with the LLM's latent space. This enables seamless integration of multimodal protein representations into the LLM, enabling information extraction from not only the protein structural and sequential information but also the rich pretrained knowledge within ESM.

\begin{figure*}[ht]
% \vskip 0.2in
\begin{center}
\centerline{\includegraphics[width=\linewidth]{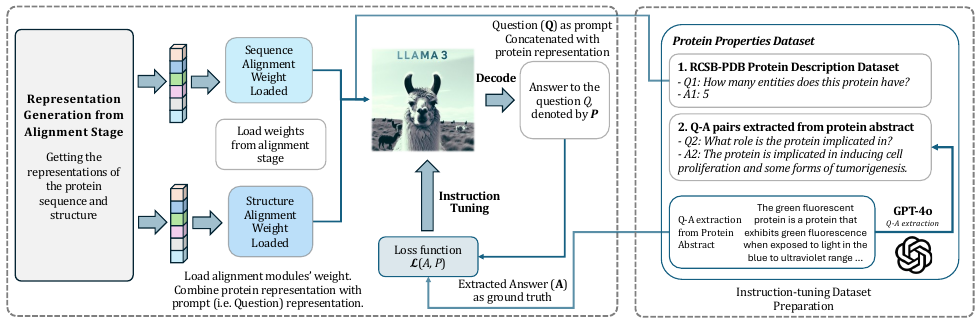}}
\vspace{-3mm}
\caption{\raisebox{-1ex}{\includegraphics[height=24pt]{figures/PGPT.png}}\hspace{-0.1em} \model Instruction Tuning Stage: we utilize the QA pairs and property tags in \dataset to tune the LLM to follow instructions and give concise responses. For instruction alignment, explicit prompts (\textit{Questions} on the protein) are included at the beginning of the prompt.}
\label{fig:schema_it}
\end{center}
\vspace{-6mm}
\end{figure*}

To train \model for effective modality alignment, we introduce \dataset, a large-scale dataset of over \datasize protein sequences with structural information and annotations. 
% Then, a projection layer aligns both embeddings with the LLM. To train the system for modality alignment, a dataset of over \datasize protein sequences, structures, and annotations is processed and used to train \model. 
Unlike previous works that use entire protein annotations as prediction targets for instruction-tuning, we use GPT-4o~\citep{openai2024gpt4} to systematically decompose proteins' abstract summary from the RCSB Protein Data Bank (RCSB PDB)~\citep{burley2023rcsb} into question-answer (QA) pairs. We then finetune \model on \dataset using diverse open-source models, including Llama-2~\citep{touvron2023llama}, Llama-3~\citep{dubey2024llama} Vicuna~\citep{vicuna}, and Mistral~\citep{jiang2023mistral}. The training effectively enhances the model's ability to understand user queries and generate concise, contextually relevant answers. Our contributions are as follows:
\vspace{-2mm}
\begin{itemize}[leftmargin=1em]
    \item \textbf{Novel Framework.} We introduce \model, a state-of-the-art protein LLM that fuses protein sequence and structural information to enable interactive protein-focused conversations, significantly enhancing the understanding and design of proteins;
    \item \textbf{Large-scale Dataset.} We propose \dataset, a large-scale protein dataset based on  RCSB-PDB~\citep{guo2023proteinchat}. \dataset encompasses 132,092 protein samples, each annotated with a detailed descriptive abstract, 20-30 property tags, and 5-10 QA pairs. The depth and variety of these annotations position \dataset as a high-quality instruction tuning corpus;
    \item \textbf{Comprehensive Experiments.} We conducted extensive experiments on mainstream open-source and proprietary LLM backbones under different scenarios. Our empirical analysis provides guidance for future design of protein LLMs. % for better alignment.
\end{itemize}

% \begin{itemize}[leftmargin=1em]
%     \item \textbf{Modality Fusion:} 
%     \item \textbf{Prior Knowledge Incorporation:} We propose the usage of ESM-2 on both modalities \footnote{Prior knowledge about protein's properties and structures is imprinted into the parameters of ESM-2 models. ESM-2's sequence model encodes knowledge on protein sequence patterns while ESM-2's folding model encodes physics laws about protein structure.}. The multi-modality fusion alignment adds additional context to \model for higher-quality response generation.
%     \item \textbf{GPT-4 QA Corpus:} We propose the decomposition of complex protein annotations specific QA pairs for better model understanding in instruction tuning. 
% \end{itemize}

\section{Methodology}

\subsection{Model Architecture}

\model consists of two frozen pre-trained encoders (Figures \ref{fig:schema_ma} and \ref{fig:schema_it}): an inverse folding model (\verb|esm_if1_gvp4_t16_142M_UR50|) for structure encoding and a protein language model for sequence encoding (\verb|esm2_t36_3B_UR50D|). 
The embeddings generated by these models are fed into a linear projection layer to produce soft prompts for the LLM. The model training comprises two stages: 1) {\it Sequential and Structural Alignment} and 2) {\it Instruction-tuning}.
% Both embeddings are then fed into a linear projection layer to produce soft prompts for the LLM. The training of \model is split up into two stages: {\it alignment} and {\it instruction tuning}.

% \input{contents/fig/architecture}

% The training process includes 2 stages: a) modality fusion \& alignment, and b) instruction tuning.

\subsubsection{Sequence and Structure Alignment}
In the alignment stage, protein structures are first fed into the pre-trained structure encoder \verb|esm_if1_gvp4_t16_142M_UR50| which explicitly captures the detailed 3D structures and models spatial interactions between amino acid residues. Then, sequences are encoded using the sequence encoder \verb|esm2_t36_3B_UR50D| featuring 36 transformer layers and 3 billion parameters, trained on the Protein \verb|UniRef50/D| database to enhance sequence diversity.
This module integrates structural information with implicit structural contact, evolutionary, and biochemical information that the structure alone does not capture. For efficient training, both of these modules are frozen.
% protein sequence/structure as placeholders in the instruction tuning template
We utilize a specialized token prompt for protein-text modality alignment: 
$$
\begin{array}{l}
    \mathbf{Q}{:}<Protein><Struct><Seq>\\</Protein><QuestionPrompts>\\\mathbf{A}{:}<Description>
\end{array}   
$$
% The modalities are then fused for protein-text alignment using the prompt above, where 
% the soft prompts encoding the structural and sequential information are prepended to the question prompt.
The structural and sequential information is encoded into the soft prompts and prepended to the question prompt. 
In stage 1 training, the question prompt \textbf{Q} is left empty to prioritize learning the abstract description from the protein representation.

% The question prompt \textbf{Q} is empty for stage 1 training as the purpose is to learn the corresponding abstract description from the protein representation. 
The description tag is then replaced with the full annotation from RCSB-PDB~\citep{guo2023proteinchat} to train the projection layer in aligning a protein with its annotation description.

\subsubsection{Instruction-tuning}
In stage 2, the model undergoes instruction tuning using our curated QA dataset. Unlike previous works that utilize full annotations, we focus on specific QA examples to facilitate instruction tuning. 
We augment the abstract dataset from stage 1 using GPT-4o to generate explicit QA pairs for this stage. 
% IWe wrap the prompts used in stage 1 with LLaMA style $\#\#\# Human:$ and $\#\#\# Assistant:$ prompts.
The prompts from stage 1 are adapted to the LLaMA style (``\texttt{\#\#\# Human:''} $\ldots$ and \texttt{\#\#\# Assistant:} $\ldots$), with \textbf{Q} replaced by explicit questions from the QA dataset, such as ``how many assemblies does this protein have.'' The model then generates descriptive yet concise answers from the given protein as \textbf{A}.

\subsection{Dataset Curation}
\label{sec:dataset}

To ensure the highest quality of training data, we implement a rigorous validation and data collection process. 
We leverage RCSB-PDB dataset~\citep{Berman2000}, which is thoroughly vetted by both RCSB-PDB~\citep{Berman2000} and PubMed\footnote{\url{https://pubmed.ncbi.nlm.nih.gov/}} for reliable protein information. The dataset is derived from peer-reviewed PubMed publications, implying that the dataset content is highly accurate. 
We select a large dataset to cover a wide range of proteins and ensures coverage on out-of-distribution proteins. 
For modality alignment, we construct a large-scale dataset from the RCSB-PDB database~\citep{Berman2000} consisting of \datasize protein structures, sequences, and abstract descriptions. The raw dataset of 204,826 proteins is filtered to retain only those with an abstract description, chain A, and sequences without non-encodable characters. % and do not have non-encodable characters in their sequence. 
Each entry in the final dataset includes the 3D protein structure represented by backbone atomic coordinates, the sequence string, and a rich protein annotation, as shown in Figure~\ref{fig:schema_it}. The detailed statistics of our dataset are presented in Table \ref{tab:summary_stats}, highlighting the extensive annotations and comprehensive content available for each protein. 
% \yq{Do we have a table for more fine-grained dataset statistics?}
% In the final dataset, each entry contains the 3D protein structure represented in their backbone atomic coordinates, the sequence string, and a rich protein annotation as seen in Figure \ref{fig:schema_it}. 
% \yq{Details for dataset construction is in Appendix X (add details about how we use GPT-4 for data generation, as well as the prompts.)}

\begin{table}[h!]
\centering
% \vspace{-2mm}
\begin{tabular}{lccc}
\toprule
\textbf{Per Protein} & \textbf{Min} & \textbf{Max} & \textbf{Mean} \\
\midrule
\midrule
\textbf{\# Abstracts tokens} & 89 & 728 & 205.45 \\
\textbf{\# Answer tokens} & 32 & 550 & 98.56 \\
\midrule
\textbf{\# Open-Ended QA Pair samples} & 10 & 26 & 17.39 \\
\textbf{\# Closed-Ended QA Pair samples} & 24 & 29 & 25.94 \\
\bottomrule
\end{tabular}
\caption{Statistics for \dataset}
\label{tab:summary_stats}
\end{table}
\vspace{-5mm}
\subsection{Data Augmentation}
\vspace{-2mm}
Previous works often use the entire protein annotation for instruction tuning~\citep{xu2023protst, guo2023proteinchat, wang2024protchatgpt}, which may result in the model producing overly detailed responses with extraneous information not directly relevant to the user prompt.
% generating overly detailed responses containing extra information that does not directly respond to the user prompt. 
Therefore, our \model decomposes the rich protein annotations into more specific QA-pairs for instruction tuning so that user instructions can be concisely answered. We do this by generating short concise answers along with long-form responses to include our dataset. Concretely, we prompt GPT-4o to generate both open-ended and close-ended QA pairs with the context of the abstract to decompose the abstract into atom-level QA pairs. As seen in Table \ref{tab:summary_stats}, on average, each protein has around 40 total QA pairs generated from this process.

\vspace{-2mm}
\section{Experiments}
\vspace{-2mm}
\subsection{Training}
We trained \model on 4 base LLM architectures: Vicuna~\citep{vicuna}, LLaMA-2~\citep{zhang2023llama}, LLaMA-3~\citep{dubey2024llama}, and Mistral~\citep{jiang2023mistral}. 
Our training process is divided into two phases: modality alignment (MA) and instruction tuning (IT). This approach allows the model to preserve previously acquired knowledge while effectively handling specific instructions, such as protein-related queries. 

\ifthenelse{\boolean{forKDD}}{
\input{tables/tab-lexical}
\begin{table*}[htbp]
    \centering
    % \begin{minipage}{.98\linewidth}{
    \setlength{\tabcolsep}{3.2pt} 
    \begin{tabular}{ll|rrr|rrr|rrr}
    \toprule
    \multirow{2}{*}{Metric} & Base & \multicolumn{3}{c|}{(a) Protein Sequence} & \multicolumn{3}{c|}{(b) Modality Fusion} & \multicolumn{3}{c}{(c) \model} \\
     & Model & $S_{\mathrm{BERT}}$ & $S_{\mathrm{Pub}}$ & $S_{\mathrm{GPT}}$ & $S_{\mathrm{BERT}}$ & $S_{\mathrm{Pub}}$ & $S_{\mathrm{GPT}}$ & $S_{\mathrm{BERT}}$ & $S_{\mathrm{Pub}}$ & $S_{\mathrm{GPT}}$ \\
    \midrule
    \multirow{4}{*}{\textbf{Pre}} 
    & Vicuna & 0.572 & 0.464  & 0.396 &  0.582 & 0.515 & 0.446 & 0.730 & 0.626 & 0.605 \\
    & Llama-2 & 0.513 & 0.372 & 0.362 & 0.589 & 0.446 & 0.414 & 0.739 & 0.644 & 0.606 \\
    & Llama-3 & 0.490 & 0.442 & 0.369 & 0.593 & 0.487 & 0.446 & 0.779 & 0.706 & 0.698 \\
    & Mistral & 0.525 & 0.405 & 0.362 & 0.513 & 0.479 & 0.400 & 0.821 & 0.758 & 0.717 \\ 
    \midrule
    \multirow{4}{*}{\textbf{Rec}} 
    & Vicuna & 0.653 & 0.473 & 0.310 & 0.691 & 0.540 & 0.334 & 0.788 & 0.751 & 0.661 \\
    & Llama-2 & 0.680 & 0.324 & 0.214 & 0.679 & 0.477 & 0.308 & 0.796 & 0.773 & 0.689 \\
    & Llama-3 & 0.657 & 0.332 & 0.210 & 0.695 & 0.456 & 0.309 & 0.803 & 0.776 & 0.713 \\
    & Mistral & 0.624 & 0.287 & 0.192 &  0.623 & 0.359 & 0.211 & 0.839 & 0.816 & 0.752 \\
    \midrule
    \multirow{4}{*}{\textbf{F1}} 
    & Vicuna & 0.608 & 0.468 & 0.347 & 0.629 & 0.524 & 0.381 & 0.756 & 0.682 & 0.630 \\
    & Llama-2 & 0.582 & 0.345 & 0.269 & 0.628 & 0.459 & 0.351 & 0.764 & 0.701 & 0.644 \\
    & Llama-3 & 0.560 & 0.378 & 0.268 & 0.638 & 0.470 & 0.363 & 0.790 & 0.739 & 0.705 \\
    & Mistral & 0.569 & 0.335 & 0.250 & 0.561 & 0.409 & 0.276 & 0.829 & 0.784 & 0.733 \\
    \bottomrule
    \end{tabular}
    \caption{Semantic Performance of LLM with Protein Sequence as Text Input (left), with Modality Fusion ONLY (middle), and with \model (right). $S_{\mathrm{BERT}}$, $S_{\mathrm{Pub}}$, and $S_{\mathrm{GPT}}$ stand for BERTScore~\cite{zhangbertscore}, PubMedBERT-Score~\cite{pubmedbert}, and GPT-4o~\cite{openai2024gpt4} score, respectively.}
    \label{tab:semantic}
\end{table*}
\begin{table*}[htbp]
    \centering
    \setlength{\tabcolsep}{3.2pt}
    \begin{tabular}{lcccc}
    \toprule
    \textbf{Model} & \textbf{R-1} & \textbf{R-2} & \textbf{R-L} & \textbf{R-Lsum} \\
    \midrule
    \textbf{GPT-3.5-turbo} & 0.137 & 0.010 & 0.140 & 0.140 \\
    \textbf{GPT-4-turbo} & 0.049 & 0.001 & 0.049 & 0.048 \\
    \textbf{GPT-4o} & 0.068 & 0.000 & 0.067 & 0.068 \\
    \midrule
    \textbf{OpenAI o1-mini} & 0.041 & 0.001 & 0.040 & 0.040 \\
    \textbf{OpenAI o3-mini} & 0.072 & 0.007 & 0.072 & 0.073 \\
    \textbf{Deepseek-R1} & 0.003 & 0.000 & 0.003 & 0.003 \\
    \midrule
    \textbf{ProteinGPT\textsubscript{Vicuna}} & 0.345 & 0.007 & 0.350 & 0.348 \\
    \textbf{ProteinGPT\textsubscript{Llama-2}} & \underline{\textit{0.348}} & 0.014 & \underline{\textit{0.347}} & \underline{\textit{0.349}} \\
    \textbf{ProteinGPT\textsubscript{Llama-3}} & 0.366 & \underline{\textit{0.021}} & 0.368 & 0.367 \\
    \textbf{ProteinGPT\textsubscript{Mistral}} & \textbf{0.461} & \textbf{0.048} & \textbf{0.460} & \textbf{0.457} \\
    \bottomrule
    \end{tabular}
    \caption{Lexical Performance with \model, OpenAI's GPT/o-series models and DeepSeek R1 model. OpenAI and DeepSeek's models are text-based models. Therefore, protein sequences are fed into the model in text format, as part of the prompt.}
    \label{tab:compare_rouge}\end{table*}

\begin{table*}[htbp]
\centering
% \small
\caption{Lexical Performance of LLM with Protein Sequence as Text Input ONLY (Left), Modality Alignment ONLY (Middle), and \model (Right). R-1, R-2, R-L, R-Lsum stand for ROUGE-1, ROUGE-2, ROUGE-L, ROUGE-Lsum, respectively.}
% \begin{tabular}{l|rrrr|rrrr|rrrr}
{\setlength{\tabcolsep}{3.5pt} 
\begin{tabular}{l|cccc|cccc|cccc}
\toprule
& \multicolumn{4}{c|}{(a) Protein Sequence} & \multicolumn{4}{c|}{(b) Modality Fusion} & \multicolumn{4}{c}{(c) \model} \\
\textbf{Model} & R-1 & R-2 & R-L & R-Lsum & R-1 & R-2 & R-L & R-Lsum & R-1 & R-2 & R-L & R-Lsum \\
\midrule
\textbf{Vicuna} & 0.026 & 0.034 & 0.025 & 0.025 & 0.104 & 0.003 & 0.105 & 0.104 & 0.345 & 0.007 & 0.350 & 0.348 \\
\textbf{Llama-2} & 0.010 & 0.001 & 0.010 & 0.010 & 0.063 & 0.004 & 0.064 & 0.064 & 0.348 & 0.014 & 0.347 & 0.349 \\
\textbf{Llama-3} & 0.005 & 0.000 & 0.005 & 0.005 & 0.090 & 0.005 & 0.090 & 0.090 & 0.366 & 0.021 & 0.368 & 0.367 \\
\textbf{Mistral} & 0.007 & 0.001 & 0.007 & 0.007 & 0.008 & 0.002 & 0.008 & 0.008 & 0.461 & 0.048 & 0.460 & 0.457 \\
\bottomrule
\end{tabular}
}
\label{tab:rouge}
\end{table*}

}{}

\noindent \textbf{Stage I: Modality Fusion/Alignment (MA).} 
In this stage, we focus exclusively on training the projection adapter by freezing both sequence and structure encoders. We set the maximum text length of abstracts to 384 characters to accommodate the annotation lengths within the RCSB-PDB dataset. The projection layer is trained over 10 epochs with a batch size of 1, weight decay of $0.05$, and 2048 warm-up steps. The dataset is divided into a training set (70\%) of 105,673 proteins and a testing set (30\%) of 26,419 proteins. We utilize the AdamW optimizer with $\beta_1=0.9, \beta_2=0.98$ \citep{loshchilov2019decoupled}, and employ a learning rate scheduler with a linear warm-up followed by cosine annealing. We set the initial learning rate to $1 \times 10^{-4}$, the minimum learning rate to $8 \times 10^{-5}$, and the warm-up learning rate to $1 \times 10^{-6}$. Automatic mixed precision (AMP)~\citep{micikevicius2018mixed} was used to improve training efficiency.

\noindent \textbf{Stage II: Instruction Tuning (IT).}
In this stage, the model is fine-tuned on a protein question-answering task. Training is conducted 
% focusing on protein question-answering instruction tuning, 
for 10 epochs with a batch size of 1, weight decay of $0.05$, and 200 warm-up steps. The QA dataset used in this phase comprises approximately 3.7 million samples, with around 35 questions per protein. %culminating in a total training dataset of about $105,673 \times 35 = 3,698,555$. 
We apply similar settings for the AdamW optimizer and AMP, but with a lower initial learning rate of $1 \times 10^{-5}$, a minimum rate of $1 \times 10^{-6}$, and a warm-up rate of $1 \times 10^{-6}$.

\vspace{-2mm}
\subsection{Inference}
\vspace{-2mm}
% For \textbf{missing or incomplete sequences}, we use the \verb|ESM-IF1| inverse folding model to predict absent segments, generating protein sequences from backbone atom coordinates. 

% For \textbf{missing structural data}, we use \verb|ESMFold| to predict the 3D structures of the missing regions, predicting 3D conformations without requiring multiple sequence alignments.
In real-world scenarios, there are cases where only protein sequence or structure information is available. Under such scenarios, protein folding and protein inverse-folding models are applied to obtain the missing structure or sequence information.

For \textbf{missing or incomplete sequences}, we use the \verb|ESM-IF1| inverse folding model to predict absent segments, generating protein sequences from backbone atom coordinates. This method is effective in filling sequence gaps where traditional approaches struggle.

For \textbf{missing structural data}, we apply \verb|ESMFold| to predict 3D structures of missing regions, generating accurate 3D conformations without relying on multiple sequence alignments.

% For \textbf{missing or incomplete sequences}, we leverage the \verb|ESM-IF1| inverse folding model to predict the absent segments in the protein structure. This model is capable of generating protein sequences directly from backbone atom coordinates, which allows for the reconstruction of missing parts of a protein sequence based on the spatial configuration of its known elements. By applying this technique, we can predict the sequence of missing segments with a high degree of accuracy, ensuring that the reconstructed sequence maintains the overall structural integrity of the protein.

% For \textbf{missing structural data}, we utilize the \verb|ESMFold| model to predict the 3D structures of the regions with unavailable structural information. \verb|ESMFold| is designed to predict 3D conformations of protein regions without the need for multiple sequence alignments, which is particularly beneficial in cases where sequence alignment data is sparse or unavailable. This model relies on the understanding of protein folding mechanisms and can generate accurate structural predictions for missing regions, thereby enhancing the completeness and accuracy of protein structure models.

% For missing sequences, we use the inverse folding ESM model to generate them, and for missing structural data, the forward folding ESM model predicts 3D structures. This ensures reliable outputs even with partial data.

\vspace{-2mm}
\subsection{Computational Cost} 
\vspace{-2mm}
Our training uses two NVIDIA H100 PCIe GPUs (80GB vRAM) and two NVIDIA A100 PCIe GPUs (40GB). We implemented strategies including Automatic mixed precision (AMP), optimized data loaders, asynchronous data processing, multi-GPU training. Stage 1 and 2 require approximately one week and 60 hours, respectively.

\model is optimized for fast processing, handling user protein queries in $\sim50$ seconds and generating 23-token responses in about 8 seconds. Its computational efficiency allows deployment on hardware ranging from GPUs like the NVIDIA T4 with 16 GB memory to standard CPUs. 

\vspace{-3mm}
\section{Results}
\vspace{-3mm}

\begin{figure}[htbp]
\centering
\raisebox{7mm}
{
\begin{subfigure}{0.49\linewidth}
    \includegraphics[width=0.98\linewidth]{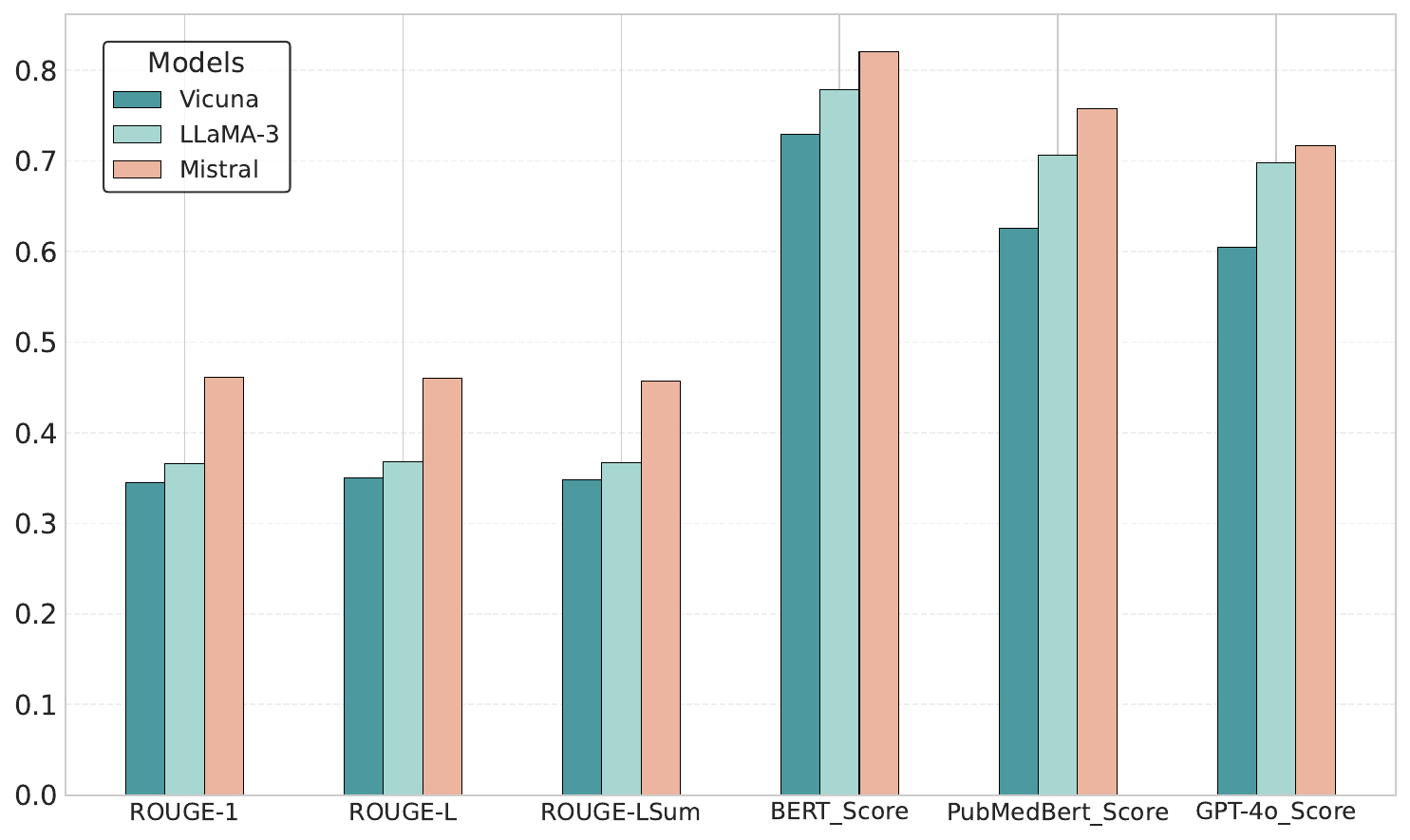}
\caption{\model performance with various base LLMs.}
\label{fig:vis_proteingpt}
\end{subfigure}
}
\begin{subfigure}{0.49\linewidth}
\includegraphics[width=0.98\linewidth]{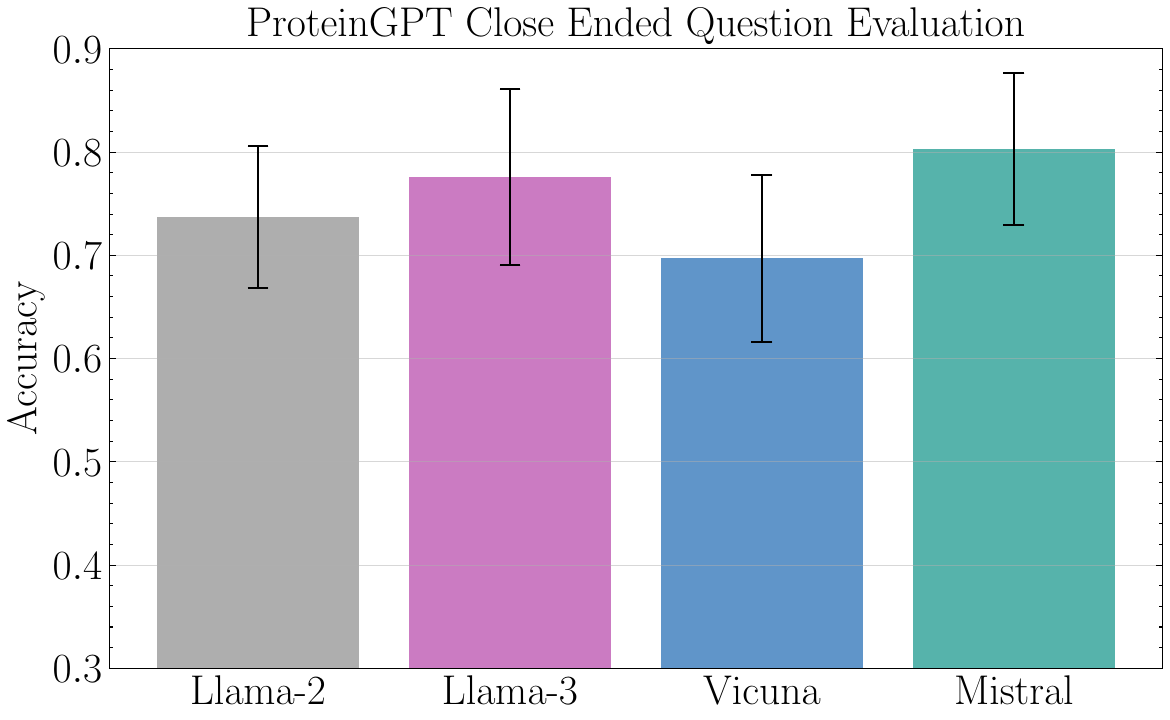}
\caption{\model performance (after instruction-tuning) on fact-based, closed-ended questions, such as determining number of polymer entities in a given protein.}
\label{fig:close_acc}
\end{subfigure}
\vspace{-8mm}
\end{figure}

% \begin{figure}[htbp]
% \centering
% \vspace{-2mm}
% \caption{\model performance (after instruction-tuning) on fact-based, closed-ended questions, such as determining number of polymer entities that exist in a given protein.}
% \label{fig:close_acc}
% \vspace{-5mm}
% \end{figure}

We conducted a series of experiments to assess \model's effectiveness both quantitatively and qualitatively. 
% Additionally, we carried out ablation studies to ascertain the contributions and importance of various modules at different stages. These included the original model, the modality-aligned model, and the instruction-tuned version. 
Moreover, we benchmark the capabilities of state-of-the-art large language models on protein-related tasks. % Instances of empty model outputs are substituted with ``N/A'' for data consistency.

\subsection{Qualitative Evaluation}
\begin{figure*}[ht]
% \vskip 0.2in
\begin{center}
\centerline{\includegraphics[width=0.9\textwidth]{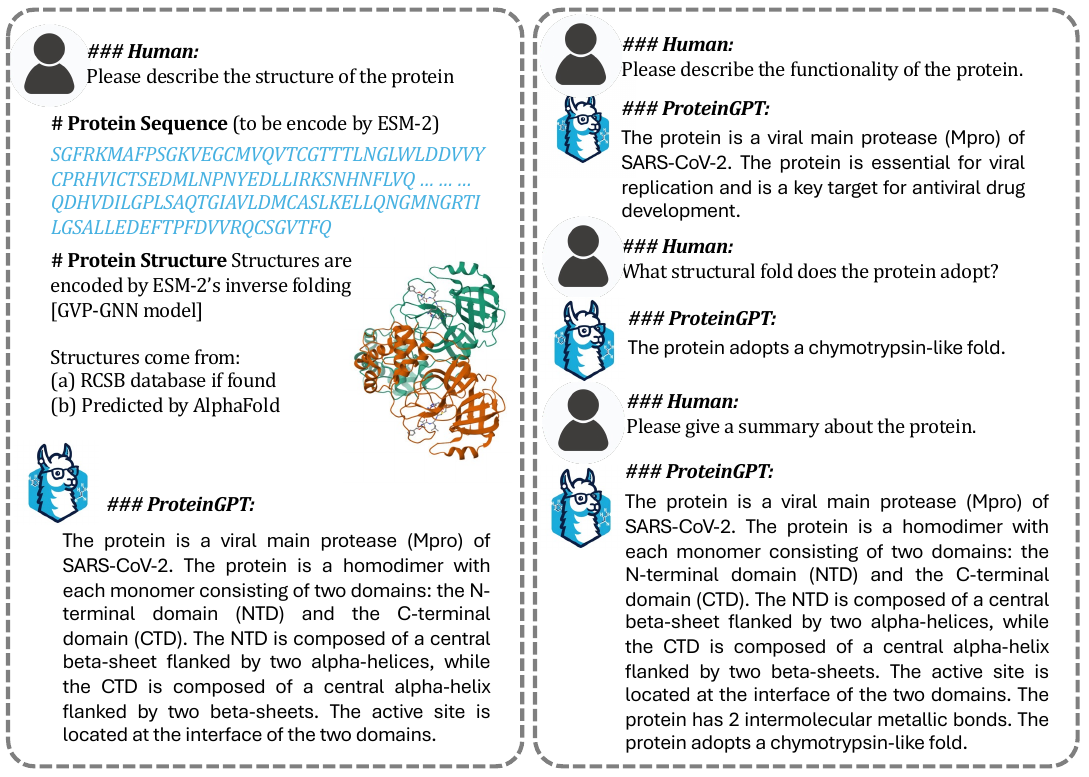}}
% \vspace{-2mm}
\caption{Conversations between humans and \model on Protein 6O7Q, where \model provides detailed insights into both sequence (e.g., \emph{60-subunit MoFe proteins}) and structural information (e.g., \emph{substrate azide and product ammonia}).}
\label{fig:6O7Q}
\end{center}
\vspace{-7mm}
\end{figure*}

Figures~\ref{fig:6O7Q}, \ref{fig:5X1Y}, \ref{fig:7RUV} show example conversations between human users and \model on Proteins 6O7Q, 5X1Y, and 7RUV, respectively. To ensure unbiased evaluation and avoid data leakage, all testing was performed on a separate set of proteins isolated from training. %which ensures that \model has never seen the proteins before. 
\model effectively interpret the semantics of queries and produces accurate, logically consistent responses. 
These responses include details about protein \emph{functions}, such as catalyzing the reduction of dinitrogen to ammonia, and \emph{structures}, such as the structural dependencies on the substrate azide and the product ammonia. This demonstrates \model's capabilities on protein sequence, structure, and function understanding tasks and its potential for enabling rapid exploration of proteins.

% in terms of both the protein's sequence information, such as correctly indicating the protein's function of catalyzes the reduction of dinitrogen to ammonia, and structural information, such as the structual dependency on the substrate azide and the product ammonia. 
% This implies that researchers can swiftly survey a protein's structure, sequences, functions, and other topics through \model. 

\vspace{-2mm}
\subsection{Quantitative Evaluation}
\vspace{-2mm}
\subsubsection{Experimental Setup} 
We evaluate our model using a curated dataset of 3,508 randomly-selected question-answer-protein pairs, covering 160 proteins from the test split. Each protein is associated with 28-30 questions that were not seen by \model during training. 
We benchmark against several baseline models, including vanilla open-source LLMs (without modality alignment or instruction tuning) and proprietary models (GPT 4o/4/3.5, o3/o1-mini, and DeepSeek-R1). To manage computational costs, we sample 1,025 questions (35 proteins with 28$\sim$30 questions per protein). 
For fairness of comparison, we employ standard metrics to compare model predictions with the ground truth. These include 1) \textbf{semantic similarity} metrics that measure contextual meanings: BERTScore ($S_{\mathrm{BERT}}$)~\citep{zhangbertscore}, PubMedBERT Score ($S_{\mathrm{Pub}}$)~\citep{pubmedbert}, and GPT Score ($S_{\mathrm{GPT}}$)~\citep{openai2023gpt4}; 2) \textbf{lexical quality} metrics that measure surface-level similarity based on n-gram overlaps: ROUGE-1/2/L~\citep{ganesan2018rouge}.  For the GPT Score, we use OpenAI's \verb|text-embedding-3-large| as the embedding  model. 

% \subsubsection{Instruction Tuning Observations} Table~\ref{tab:semantic} and~\ref{tab:rouge} present the performance of the full modality-aligned and instruction-tuned \model. \model achieves high overall semantic scores, but low lexical scores. This can be explained by looking closer at the output of \model. While our annotations often contained concise responses such as "10", \model would often output complete sentences such as "the protein has 10 assemblies". Since the ROUGE score, BLEU score, METEOR score, and other similar lexical metrics reward for overlaps and penalize for misses, these lexical differences make our scores very low \citep{ganesan2018rouge, meteor, bleu}. However, in this case, the semantic meaning outweighs the exact format in which the answer is produced, we conclude that the semantic metrics serve as a better benchmark for the performance of \model.

\vspace{-2mm}
\subsubsection{Comparison among \model Variants}
\vspace{-2mm}
Among the 4 variants, ProteinGPT\textsubscript{Mistral} performs the best in terms of both semantic and lexical metrics. In terms of BERTScore (Table~\ref{tab:semantic}), ProteinGPT\textsubscript{Mistral} achieves  0.829, followed by  ProteinGPT\textsubscript{LLaMA-3} (0.790), ProteinGPT\textsubscript{LLaMA-2} (0.764), and ProteinGPT\textsubscript{Vicuna} (0.756). The strong performance of ProteinGPT\textsubscript{Mistral} can be attributed to its integration of sliding window attention (SWA)~\citep{jiang2023mistral}. Protein sequences are inherently lengthy and complex, often requiring models to capture intricate dependencies across extended stretches of amino acid. SWA helps capture local patterns and dependencies within protein sequences, leading to a longer effective attention span crucial for tasks like secondary structure prediction and functional annotation. The reduced computational load associated with SWA allows for the processing of longer protein sequences without a proportional increase in resource consumption. Figure~\ref{fig:vis_proteingpt} shows the visual comparison among different base LLMs of \model.

% which gives it a longer effective attention span to handle long sequences more effectively with a reduced inference cost, which is important for processing long-form multiple turns of QA and long inputs like protein sequences. 
% From the 4 models that we selected, the versions of \model that are based on Mistral and LLaMA-3 appear to perform the best. This can be observed in Figure~\ref{fig:vis_proteingpt}. This is likely due to the already stellar performance of these foundational LLMs, especially with LLaMA-3 being Meta's newest and most powerful model and Mistral similarly being the SOTA released by Mistral AI.

\vspace{-2mm}
\subsubsection{Baseline Comparison}
\vspace{-2mm}
We also compare \model to three groups of baselines to demonstrate \model's effectiveness in protein-specific multimodal tasks: 1) \emph{Vanilla Open-source LLMs} (Vicuna, Mistral, LLaMA-3, and LLaMA-2); 2) \emph{Proprietary General-Purpose LLMs} (GPT-4o/4/3.5); 3) \emph{State-of-the-art models with Strong Reasoning Capabilities} (OpenAI o1/o3-mini and DeepSeek-R1). 
For fairness of comparison, we prepended the protein's FASTA sequence to the prompt to provide context for these LLMs. Tables~\ref{tab:semantic} and~\ref{tab:rouge} show model performance with semantic and lexical scores. 

\noindent \textbf{Vanilla open-source LLMs} exhibit low semantic performance. When providing protein sequences as text input (Table~\ref{tab:semantic}a), the BERTScore ($S_{\mathrm{BERT}}$) range from 0.490 (LLaMA-3) to 0.572 (Vicuna) in terms of precision, indicating a lack of protein-specific pretraining and limited semantic understanding of protein data. Meanwhile, incorporating modality fusion (where additional cues beyond the raw protein sequence are integrated) leads to modest improvements, such as an improvement from 0.572 to 0.582 when using Vicuna as the base model. Model 
% Note: These are the same results as the stage in the ablation study in which no training is done. 
Similarly, \textbf{proprietary models} such as GPT-4o/4/3.5, OpenAI o1/o3-mini, and DeepSeek-R1 also exhibit lower semantic and lexical performance when processing protein sequences as text inputs. This performance gap is likely due to these models not being pretrained on domain-specific data. 
% and GPT-3.5 as the LLM are presented in Table~\ref{tab:compare_rouge} and Table~\ref{tab:compare_semantic}. We utilized \verb|GPT-3.5-turbo|, \verb|GPT-4o|, and \verb|GPT-4-turbo| in our evaluation.

\begin{figure*}[!h]
% \vskip 0.2in

\begin{center}
\centering
\includegraphics[width=1\linewidth]{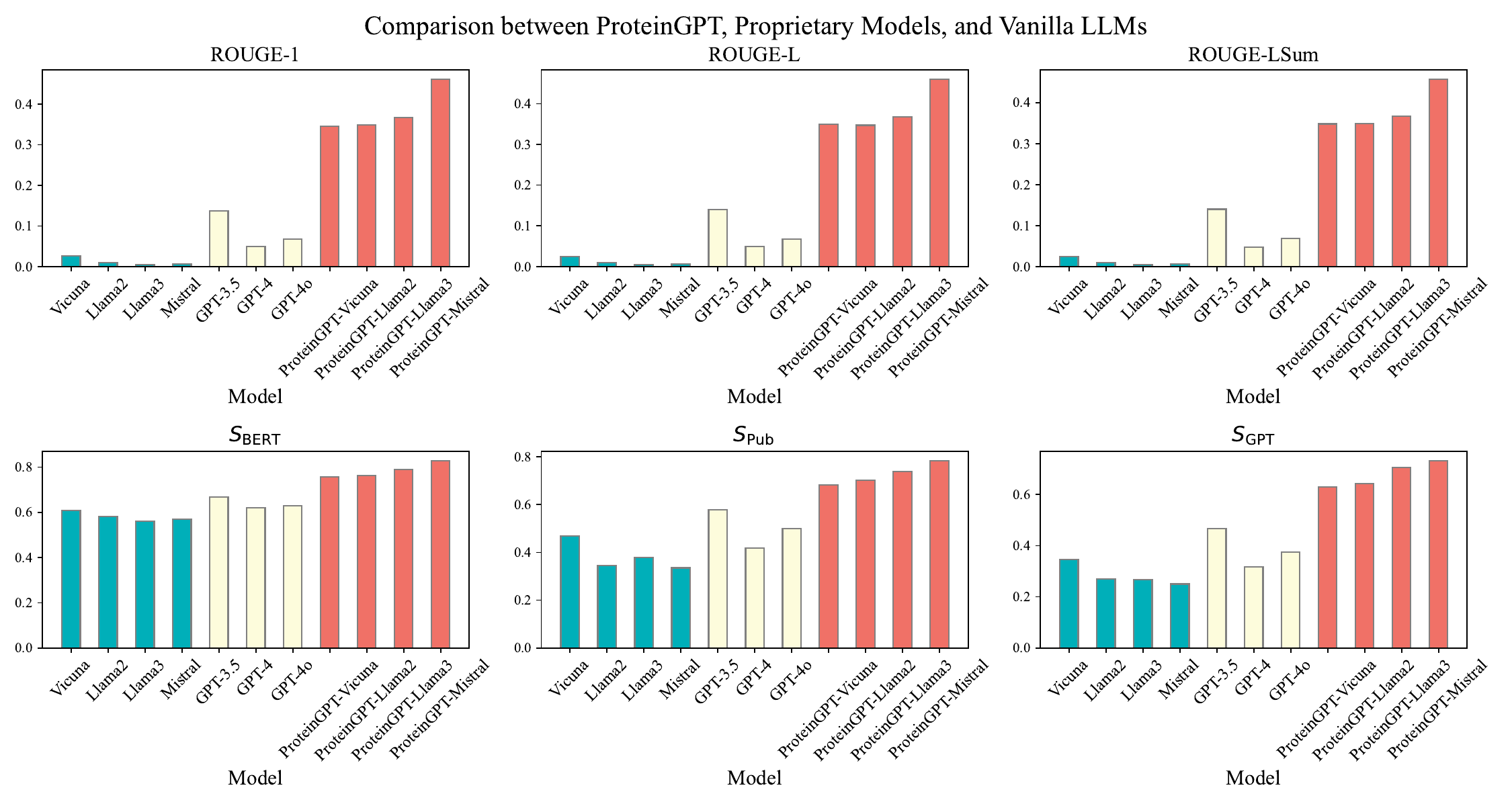}

% \caption{Comparison of ProteinGPT, OpenAI's GPT and MA models\footnote{LLM with modality alignment, but without instruction tuning.}.}
\vspace{-2mm}
\caption{Protein Text LLM takes protein primary sequence as part of the prompt to the model. GPT models are more powerful than open-source LLMs like LLaMA and Mistral. Given the same protein sequence as input, \model utilizes the information from sequence and structure encoders and yields more accurate responses.}
\label{fig:gpt_pgpt}
\end{center}
\vspace{-7mm}
\end{figure*}

The results in Table~\ref{tab:compare_rouge} highlight a critical challenge in applying general-purpose language models to protein-related queries: standard models, even those with strong reasoning capabilities like o3-mini and DeepSeek-R1, struggle to interpret protein sequences. o3-mini achieves a marginally higher Rouge-L score (0.072) than GPT-4o (0.067), but remains far below ProteinGPT\textsubscript{Mistral} (0.460). DeepSeek-R1 performs particularly poorly, which suggests that it struggles to extract meaningful insights from protein sequences when they are formatted as text. 
The sample answers in Appendix~\ref{app:deepseek_answers} show that, while models like DeepSeek-R1 demonstrates strong reasoning capabilities on general QA tasks, it struggles with domain-specific terminology such as \texttt{assemblies}, interpreting inputs merely as a \texttt{long string of amino acids}. It tends to generate verbose, speculative responses, often introducing uncertainty by stating \texttt{Maybe the question is about...}--which diminishes its utility for precise scientific queries. As a result, its performance is comparable to or slightly worse than the modality-aligned version of \model.

The overall comparison can be seen in Figure \ref{fig:gpt_pgpt}, which shows that \model outperforms both baselines consistently. This demonstrates that our model outperforms knowledge embedded within LLMs and effectively utilizes sequence and structure information to answer questions.

\textbf{Baseline Comparisons to \model} We also compare \model to two baselines to demonstrate our contributions in creating multimodal LLMs that are more capable than general-purpose LLMs in communicating about proteins. The first baseline is the vanilla LLMs that we trained our models on, such as Vicuna, Mistral, LLaMA-3, and LLaMA-2. The second baseline is GPT-4 and GPT-3.5. For evaluation, we simply pretended the FASTA sequence of the protein in front of the prompt to give the LLM context of the protein. Table~\ref{tab:compare_semantic} and Figure~\ref{fig:gpt_pgpt} provide quantitative and visual comparison on baseline LLMs with \model.

\vspace{-3mm}
\subsubsection{Close-Ended Accuracy Experiment}
Although semantic-based evaluations may be useful in gauging the feasibility of our outputs, to ensure our model is outputting factually correct information regarding a given protein, we also conduct a close-ended answer format evaluation on \model with samples from our test subset of proteins. We selected 160 proteins for evaluation but only used QA-pairs that had a factual single-word ground truth and excluded questions that had open-ended answers (e.g. "describe this protein"). Examples of such closed-ended questions are "yes"/"no" questions or information on the number of assemblies or polymers in a protein. We then use GPT-4o to directly judge the outputs of \model to the ground truth in our dataset.

The results can be seen in Figure \ref{fig:close_acc}. LLaMA-3 and Mistral are the best-performing backbone models, achieving around 80\% accuracy in answering fact-based closed-ended questions. Even the weaker models like LLaMA-2 and Vicuna achieve reasonable accuracy above 70\%. Therefore, \model not only demonstrates strong capabilities in generating feasible answers as demonstrated by our semantic evaluations, but \model also provides factually accurate answers as demonstrated by this accuracy evaluation.

\begin{figure*}[!h]
% \vskip 0.2in

\begin{center}
\centering
\includegraphics[width=1\linewidth]{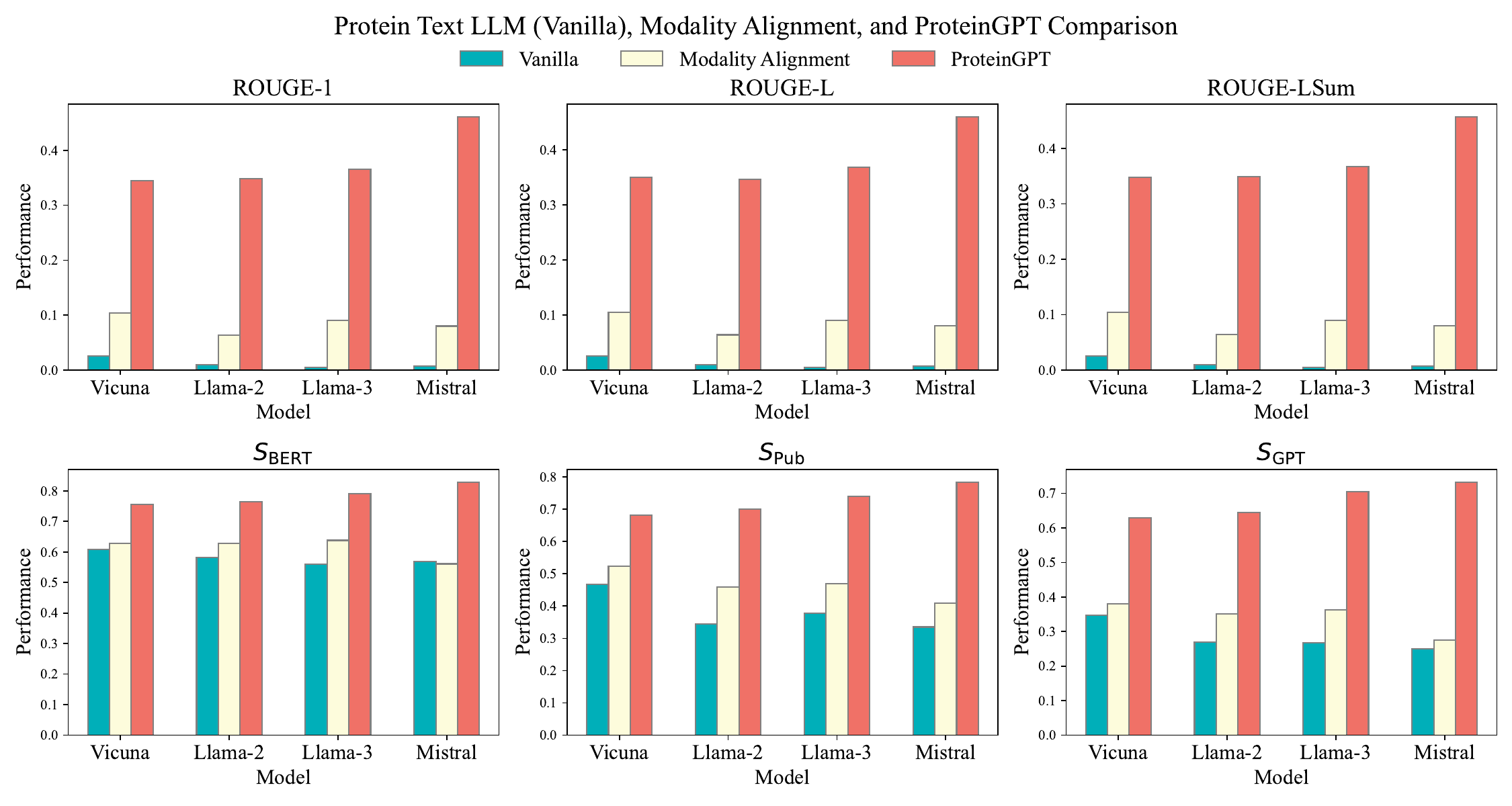}
\vspace{-5mm}
\caption{Performance improves progressively from the vanilla LLM model with protein as text to the modality-aligned version, and finally to the instruction-tuned variants of \model. Each stage of \model's training results in substantial enhancements in both lexical and semantic performance, showcasing the efficiency of our framework.}
\label{fig:perform_all}
\end{center}
\vspace{-4mm}
\end{figure*}

% \begin{figure*}[!h]
% % \vskip 0.2in

% \begin{center}
% \centering
% \includegraphics[width=1\linewidth]{figures/GPT_PGPT.pdf}

% % \caption{Comparison of ProteinGPT, OpenAI's GPT and MA models\footnote{LLM with modality alignment, but without instruction tuning.}.}
% \vspace{-4mm}
% \caption{Protein Text LLM takes protein primary sequence as part of the prompt to the model. GPT models are more powerful than open-source LLMs like LLaMA and Mistral. Given the same protein sequence as input, \model utilizes the information from sequence and structure encoders and yields more accurate responses.}
% \label{fig:gpt_pgpt}
% \end{center}
% \vspace{-4mm}
% \end{figure*}

\vspace{-3mm}
\subsection{Ablation Study}
Lastly, to better understand our contributions and ensure the model is learning at each step in the architecture, we perform an ablation study to demonstrate that the module after each stage is indeed improving the performance of \model. \model is broken down into three modules: vanilla LLM, modality-aligned LLM, and instruction-tuned LLM.

\noindent \textbf{Vanilla LLM: }Table~\ref{tab:semantic} (a) and Table~\ref{tab:rouge} (a) display the semantic and lexical scoring using similar metrics for the vanilla LLM of choice (Vicuna, LLaMA, etc.). This is evaluated on the same set of 160 proteins and 3508 questions that we used to evaluate the final model. Also note that at this stage, no training has been done, the model is the same as the out-of-the-box LLM.

% \begin{table*}[!h]
%   \centering
%   \resizebox{0.5\textwidth}{!}{
%     \begin{tabular}{lrrrr}
%     \toprule
%     \textbf{Model} & \textbf{ROUGE-1} & \textbf{ROUGE-2} & \textbf{ROUGE-L} & \textbf{ROUGE-LSum} \\
%     \midrule
%     \textbf{Vicuna} & 0.092 & 0.003 & 0.090 & 0.092 \\
%     \textbf{LLaMA-2} & 0.063 & 0.004 & 0.064 & 0.064 \\
%     \textbf{LLaMA-3} & 0.090 & 0.005 & 0.090 & 0.090 \\
%     \textbf{Mistral} & 0.008 & 0.002 & 0.008 & 0.008 \\
%     \bottomrule
%     \end{tabular}%
%     }
%   \caption{Lexical Performance with only Modality Alignment.}
%   \label{tab:ma_rouge}%
% \end{table*}%

\noindent \textbf{Modality Aligned (MA) LLM: }Following this, Table~\ref{tab:semantic} (b) and Table~\ref{tab:rouge} (b) show that of the LLM after modality alignment. Evaluated on the same set of proteins, at this stage, the linear layer has been trained to learn to align and fuse the structure and sequence modalities to the LLM.

\noindent\textbf{Instruction Tuned (IT) LLM: }Lastly, as mentioned previously, Tables \ref{tab:semantic} (c) and \ref{tab:rouge} (c) are for the fully aligned and instruction-tuned model. At this stage, the model is complete and has been tuned on our GPT-4o curated dataset to follow instructions concisely.

\begin{table*}[!h]
  \centering
  {
  \resizebox{\linewidth}{!}{
  \setlength{\tabcolsep}{4pt}
    \begin{tabular}{ll|ccc|cccc|ccc}
    \toprule
    \multicolumn{2}{l}{Scoring Metrics} & \textbf{GPT-3.5} & \textbf{GPT-4} & \textbf{GPT-4o} & \textbf{P-GPT\textsubscript{Vicuna}} & \textbf{P-GPT\textsubscript{Llama-2}} & \textbf{P-GPT\textsubscript{Llama-3}} & \textbf{P-GPT\textsubscript{Mistral}} & \textbf{o1-mini} & \textbf{o3-mini} & \textbf{DeepSeek-r1} \\
    \midrule
    \multirow{3}{*}{\textbf{S}\textsubscript{BERT}} 
        & Precision & 0.641 & 0.578 & 0.596 & 0.730 & 0.739 & 0.779 & \textbf{0.821} & 0.572 & \underline{\textit{0.597}} & 0.468 \\
        & Recall & 0.701 & 0.675 & 0.668 & 0.788 & 0.796 & 0.803 & \textbf{0.839} & 0.664 & \underline{\textit{0.699}} & 0.635 \\
        & F1 & 0.667 & 0.621 & 0.628 & 0.756 & 0.764 & 0.790 & \textbf{0.829} & 0.612 & \underline{\textit{0.641}} & 0.537 \\
    \midrule
    \multirow{3}{*}{\textbf{S}\textsubscript{Pub}} 
        & Precision & 0.513 & 0.434 & 0.440 & 0.626 & 0.644 & 0.706 & \textbf{0.758} & 0.402 & \underline{\textit{0.450}} & 0.449 \\
        & Recall & 0.667 & 0.406 & 0.580 & 0.751 & 0.773 & 0.776 & \textbf{0.816} & 0.393 & \underline{\textit{0.513}} & 0.307 \\
        & F1 & 0.579 & 0.418 & 0.499 & 0.682 & 0.701 & 0.739 & \textbf{0.784} & 0.397 & \underline{\textit{0.477}} & 0.363 \\
    \midrule
    \multirow{3}{*}{\textbf{S}\textsubscript{GPT}} 
        & Precision & 0.470 & 0.391 & 0.391 & 0.605 & 0.606 & 0.698 & \textbf{0.717} & 0.387 & \underline{\textit{0.410}} & 0.360 \\
        & Recall & 0.466 & 0.266 & 0.363 & 0.661 & 0.689 & 0.713 & \textbf{0.752} & 0.281 & \underline{\textit{0.396}} & 0.247 \\
        & F1 & 0.467 & 0.316 & 0.376 & 0.630 & 0.644 & 0.705 & \textbf{0.733} & 0.325 & \underline{\textit{0.398}} & 0.293 \\
    \bottomrule
    \end{tabular}
    }
  }
  \caption{Semantic Performance on OpenAI GPT Protein Text LLMs (GPT-3.5, GPT-4, and GPT-4o) 
  and our \model \textsubscript{\{Vicuna, Llama-2, Llama-3, Mistral\}} models in terms of BERTScore ($S_{\mathrm{BERT}}$)~\citep{zhangbertscore}, PubMedBERT Score ($S_{\mathrm{Pub}}$)~\citep{pubmedbert}, and GPT-4o Score ($S_{\mathrm{GPT}}$)~\citep{openai2024gpt4}.}
  \label{tab:compare_semantic}%
  \vspace{-4mm}
\end{table*}%

Figure~\ref{fig:perform_all} highlights the differences in performance between each of these stages. We can observe that consistently, the instruction-tuned and modality-aligned final model outperforms the modality-only model and vanilla LLMs. This falls in line with our hypothesis and demonstrates that our 3 stages of training are indeed improving the model's multi-modal understanding of proteins. More specifically, the observation that modality alignment always performs better than a vanilla LLM demonstrates that through this stage, the MLLM understands how to digest multi-modal information. Similarly, the observation that the instruction-tuned and modality-aligned \model performs better than all other stages demonstrates that this stage indeed teaches the model how to properly answer questions related to these structures and sequences it learned from the previous stage.

\vspace{-2mm}
\section{Related Works}
\vspace{-2mm}

\textbf{Protein Representation Learning.}
% Proteins, the essential functional units within cells, exhibit four hierarchical structural levels critical for their biological roles. 
Studies in protein representation mainly focus on the four hierarchical structural levels of proteins, which are critical for their biological roles. 
% features from these structural levels. 
Some research~\citep{madani2023large,notin2022tranception} treat protein sequences as a biological language and utilize the Transformer architecture \citep{transformer} to model amino acid interactions and predict sequences using large sequence databases. 
Other approaches~\citep{ESM-2,ESM-1v,ESM-1b,ESM,vig2020bertology} employ Masked Language Modeling (MLM) to develop attention mechanisms reflecting protein spatial interaction maps. 
Structure-oriented methods~\citep{gligorijevic2021structure,GVP,GearNet} encapsulate the functional attributes and spatial data of proteins for tasks like molecule binding~\citep{jin2021iterative, kong2022conditional}, protein interface studies~\citep{mahbub2022egret, reau2023deeprank}, and property predictions~\citep{zhang2022ontoprotein}. 
However, most works rely on single-modal data, which overlooks the cross-modality interactions among text and protein sequence \& structure information. 
% Although ProtChatGPT~\citep{wang2024protchatgpt} incorporates both modalities, its PLP-former alignment on full abstracts introduces complexity in the model and verbosity in responses.
% Our research employs protein language models to generate high-quality embeddings of protein sequences and structures without modifying their underlying architectures.

\noindent \textbf{Large Language Models.}
Recent advancements in Large Language Models (LLMs) such as GPT-4 \citep{openai2023gpt4}, LLaMA~\citep{touvron2023llama}, Mistral Large 2~\citep{jiang2023mistral}, and Gemini~\citep{team2023gemini}
% , and Multimodal Large Language Models (MLLMs) such as ChatGPT-4 \citep{openai2024gpt4}, Mini-GPT4 \citep{zhu2023minigpt4}, LLaVA \citep{liu2023visual}, and PaLM-E \citep{driess2023palme} 
have established new benchmarks in natural language processing (NLP), offering enhanced language comprehension and reasoning~\citep{zellers2019hellaswag, wang2019glue, frohberg2022crass, yao2024benchmarking, zhang2023benchmarking}. 
Multimodal LLMs (MLLMs) have further extended these capabilities beyond text, enabling the processing of natural language task performance on multimodal data~\citep{shu2023audiovisual, yu2023mmvet, ning2023videobench}. 
% Notably, BLIP-2\citep{BLIP2} uses a specific transformer for visual-textual alignment. FROMAGe~\citep{FROMAGe} and MedVQA~\citep{MedVQA} refine cross-modal integration, enhancing medical visual question answering and scientific comprehension with diverse biological data.
% Models like Flamingo\citep{AlayracDLMBHLMM22} and BLIP-2\citep{0008LSH23} employ advanced architectures for effective visual linkage, while LLaVA\citep{abs-2304-08485} directly maps image features into text embeddings. Advances also encompass video and audio processing with technologies like Video-Chat\citep{abs-2305-06355} and SpeechGPT\citep{abs-2305-11000}, with PandaGPT\citep{abs-2305-16355} notably handling inputs from multiple modalities.
As proteins can be naturally represented by character strings, LLMs like ProteinChat \citep{guo2023proteinchat} and ProtChatGPT \citep{wang2024protchatgpt} have been developed to effectively analyze protein structures and sequences. 

% \textbf{Instruction Tuning.} The adaptation of large language models (LLMs) such as GPT-3~\citep{brown2020language}, T5~\citep{raffel2020exploring}, PaLM~\citep{chowdhery2022palm}, and OPT~\citep{zhang2022opt} to follow natural language instructions has led to the development of specialized instruction-tuned models like InstructGPT~\citep{ouyang2022training}, FLAN-T5~\citep{chung2022scaling}, and FLAN-PaLM~\citep{chung2022scaling}. These models have shown improved zero- and few-shot generalization capabilities through instruction tuning~\cite {ouyang2022training,wang2022benchmarking,wang2022self}.
% Extending instruction tuning to the multi-modality domain, Flamingo~\citep{alayrac2022flamingo} stands out for its robust zero-shot task transfer and in-context learning abilities, often compared to GPT-3's impact in NLP. Similarly, BLIP-2~\citep{li2023blip} and VLP \citep{zhou2020unified} have been developed for image-text tasks. Vision language models like LLaMA-Adapter~\citep{zhang2023llama} and LLaVA~\citep{abs-2304-08485} enable LLaMA to process image inputs, aiming to foster the development of open-source multimodal LLMs.

\vspace{-3mm}

\section{Conclusions} 
\vspace{-1mm}

We introduce \model, a protein LLM that enhances question-answering capabilities and facilitates protein understanding with concise, informative responses. \model fuses structure with sequence modalities and enables alignments to any base LLMs. 
Our results demonstrate \model’s potential for practical applications in protein understanding and design, highlighting the value of interactive protein models as dynamic research tools. 
Looking ahead, future enhancements aim to introduce multi-user support, enabling real-time collaboration and knowledge sharing in biological research. Additionally, we are developing user-friendly interfaces and integration with existing lab workflows, ensuring effortless adoption into bioinformatics and computational biology pipelines. By bridging \model with widely used tools, we aim to drive innovation and collaboration in protein research.
% We anticipate that \model will drive innovation in protein science.

\bibliographystyle{plainnat}
\bibliography{neurips_2025}

\begin{thebibliography}{50}
\providecommand{\natexlab}[1]{#1}
\providecommand{\url}[1]{\texttt{#1}}
\expandafter\ifx\csname urlstyle\endcsname\relax
  \providecommand{\doi}[1]{doi: #1}\else
  \providecommand{\doi}{doi: \begingroup \urlstyle{rm}\Url}\fi

\bibitem[Bahdanau(2015)]{bahdanau2014neural}
Dzmitry Bahdanau.
\newblock Neural machine translation by jointly learning to align and translate.
\newblock In \emph{arXiv arXiv:1409.0473}, 2015.

\bibitem[Berman(2000)]{Berman2000}
H.~M. Berman.
\newblock The protein data bank.
\newblock \emph{Nucleic Acids Research}, 28\penalty0 (1):\penalty0 235–242, January 2000.
\newblock ISSN 1362-4962.
\newblock \doi{10.1093/nar/28.1.235}.
\newblock URL \url{http://dx.doi.org/10.1093/nar/28.1.235}.

\bibitem[Burley et~al.(2023)Burley, Bhikadiya, Bi, Bittrich, Chao, Chen, Craig, Crichlow, Dalenberg, Duarte, et~al.]{burley2023rcsb}
Stephen~K Burley, Charmi Bhikadiya, Chunxiao Bi, Sebastian Bittrich, Henry Chao, Li~Chen, Paul~A Craig, Gregg~V Crichlow, Kenneth Dalenberg, Jose~M Duarte, et~al.
\newblock Rcsb protein data bank (rcsb. org): delivery of experimentally-determined pdb structures alongside one million computed structure models of proteins from artificial intelligence/machine learning.
\newblock \emph{Nucleic acids research}, 51\penalty0 (D1):\penalty0 D488--D508, 2023.

\bibitem[Chiang et~al.(2023)Chiang, Li, Lin, Sheng, Wu, Zhang, Zheng, Zhuang, Zhuang, Gonzalez, Stoica, and Xing]{vicuna}
Wei-Lin Chiang, Zhuohan Li, Zi~Lin, Ying Sheng, Zhanghao Wu, Hao Zhang, Lianmin Zheng, Siyuan Zhuang, Yonghao Zhuang, Joseph~E. Gonzalez, Ion Stoica, and Eric~P. Xing.
\newblock Vicuna: An open-source chatbot impressing gpt-4 with 90\%* chatgpt quality, March 2023.
\newblock URL \url{https://lmsys.org/blog/2023-03-30-vicuna/}.

\bibitem[Dubey et~al.(2024)Dubey, Jauhri, Pandey, Kadian, Al-Dahle, Letman, Mathur, Schelten, Yang, Fan, et~al.]{dubey2024llama}
Abhimanyu Dubey, Abhinav Jauhri, Abhinav Pandey, Abhishek Kadian, Ahmad Al-Dahle, Aiesha Letman, Akhil Mathur, Alan Schelten, Amy Yang, Angela Fan, et~al.
\newblock The llama 3 herd of models.
\newblock \emph{arXiv:2407.21783}, 2024.

\bibitem[Frohberg and Binder(2022)]{frohberg2022crass}
Jörg Frohberg and Frank Binder.
\newblock Crass: A novel data set and benchmark to test counterfactual reasoning of large language models, 2022.

\bibitem[Ganesan(2018)]{ganesan2018rouge}
Kavita Ganesan.
\newblock Rouge 2.0: Updated and improved measures for evaluation of summarization tasks.
\newblock \emph{arXiv preprint arXiv:1803.01937}, 2018.

\bibitem[Gligorijevi{'c} et~al.(2021)Gligorijevi{'c}, Renfrew, Kosciolek, Leman, Berenberg, Vatanen, Chandler, Taylor, Fisk, Vlamakis, et~al.]{gligorijevic2021structure}
Vladimir Gligorijevi{'c}, P~Douglas Renfrew, Tomasz Kosciolek, Julia~Koehler Leman, Daniel Berenberg, Tommi Vatanen, Chris Chandler, Bryn~C Taylor, Ian~M Fisk, Hera Vlamakis, et~al.
\newblock Structure-based protein function prediction using graph convolutional networks.
\newblock \emph{Nature communications}, 12\penalty0 (1):\penalty0 3168, 2021.

\bibitem[Gu et~al.(2021)Gu, Tinn, Cheng, Lucas, Usuyama, Liu, Naumann, Gao, and Poon]{pubmedbert}
Yu~Gu, Robert Tinn, Hao Cheng, Michael Lucas, Naoto Usuyama, Xiaodong Liu, Tristan Naumann, Jianfeng Gao, and Hoifung Poon.
\newblock Domain-specific language model pretraining for biomedical natural language processing.
\newblock \emph{ACM Transactions on Computing for Healthcare}, 3\penalty0 (1):\penalty0 1–23, October 2021.
\newblock ISSN 2637-8051.
\newblock \doi{10.1145/3458754}.
\newblock URL \url{http://dx.doi.org/10.1145/3458754}.

\bibitem[Guo et~al.(2023)Guo, Huo, and Xie]{guo2023proteinchat}
Han Guo, Mingjia Huo, and Pengtao Xie.
\newblock Proteinchat: Towards enabling chatgpt-like capabilities on protein 3d structures.
\newblock 2023.

\bibitem[Jiang et~al.(2023)Jiang, Sablayrolles, Mensch, Bamford, Chaplot, Casas, Bressand, Lengyel, Lample, Saulnier, et~al.]{jiang2023mistral}
Albert~Q Jiang, Alexandre Sablayrolles, Arthur Mensch, Chris Bamford, Devendra~Singh Chaplot, Diego de~las Casas, Florian Bressand, Gianna Lengyel, Guillaume Lample, Lucile Saulnier, et~al.
\newblock Mistral 7b.
\newblock \emph{arXiv:2310.06825}, 2023.

\bibitem[Jin et~al.(2021)Jin, Wohlwend, Barzilay, and Jaakkola]{jin2021iterative}
Wengong Jin, Jeremy Wohlwend, Regina Barzilay, and Tommi Jaakkola.
\newblock Iterative refinement graph neural network for antibody sequence-structure co-design.
\newblock \emph{arXiv preprint arXiv:2110.04624}, 2021.

\bibitem[Jing et~al.(2020)Jing, Eismann, Suriana, Townshend, and Dror]{GVP}
Bowen Jing, Stephan Eismann, Patricia Suriana, Raphael~JL Townshend, and Ron Dror.
\newblock Learning from protein structure with geometric vector perceptrons.
\newblock \emph{arXiv preprint arXiv:2009.01411}, 2020.

\bibitem[Jumper et~al.(2021)Jumper, Evans, Pritzel, Green, Figurnov, Ronneberger, Tunyasuvunakool, Bates, Žídek, Potapenko, Bridgland, Meyer, Kohl, Ballard, Cowie, Romera-Paredes, Nikolov, Jain, Adler, Back, Petersen, Reiman, Clancy, Zielinski, Steinegger, Pacholska, Berghammer, Bodenstein, Silver, Vinyals, Senior, Kavukcuoglu, Kohli, and Hassabis]{JumperAlphaFold2021}
John Jumper, Richard Evans, Alexander Pritzel, Tim Green, Michael Figurnov, Olaf Ronneberger, Kathryn Tunyasuvunakool, Russ Bates, Augustin Žídek, Anna Potapenko, Alex Bridgland, Clemens Meyer, Simon A.~A. Kohl, Andrew~J. Ballard, Andrew Cowie, Bernardino Romera-Paredes, Stanislav Nikolov, Rishub Jain, Jonas Adler, Trevor Back, Stig Petersen, David Reiman, Ellen Clancy, Michal Zielinski, Martin Steinegger, Michalina Pacholska, Tamas Berghammer, Sebastian Bodenstein, David Silver, Oriol Vinyals, Andrew~W. Senior, Koray Kavukcuoglu, Pushmeet Kohli, and Demis Hassabis.
\newblock Highly accurate protein structure prediction with alphafold.
\newblock \emph{Nature}, 2021.

\bibitem[Kitadai and Maruyama(2018)]{kitadai2018origins}
Norio Kitadai and Shigenori Maruyama.
\newblock Origins of building blocks of life: A review.
\newblock \emph{Geoscience Frontiers}, 9\penalty0 (4):\penalty0 1117--1153, 2018.

\bibitem[Kobsa and Saltzman()]{Kobsa_Saltzman}
Serge Kobsa and W~Mark Saltzman.
\newblock Bioengineering approaches to controlled protein delivery.
\newblock URL \url{https://www.nature.com/articles/pr2008103}.

\bibitem[Kong et~al.(2022)Kong, Huang, and Liu]{kong2022conditional}
Xiangzhe Kong, Wenbing Huang, and Yang Liu.
\newblock Conditional antibody design as 3d equivariant graph translation.
\newblock \emph{arXiv preprint arXiv:2208.06073}, 2022.

\bibitem[Lin et~al.(2023)Lin, Akin, Rao, Hie, Zhu, Lu, Smetanin, Verkuil, Kabeli, Shmueli, et~al.]{ESM-2}
Zeming Lin, Halil Akin, Roshan Rao, Brian Hie, Zhongkai Zhu, Wenting Lu, Nikita Smetanin, Robert Verkuil, Ori Kabeli, Yaniv Shmueli, et~al.
\newblock Evolutionary-scale prediction of atomic-level protein structure with a language model.
\newblock \emph{Science}, 379\penalty0 (6637):\penalty0 1123--1130, 2023.

\bibitem[Loshchilov and Hutter(2019)]{loshchilov2019decoupled}
Ilya Loshchilov and Frank Hutter.
\newblock Decoupled weight decay regularization, 2019.

\bibitem[Madani et~al.(2023)Madani, Krause, Greene, Subramanian, Mohr, Holton, Olmos~Jr, Xiong, Sun, Socher, et~al.]{madani2023large}
Ali Madani, Ben Krause, Eric~R Greene, Subu Subramanian, Benjamin~P Mohr, James~M Holton, Jose~Luis Olmos~Jr, Caiming Xiong, Zachary~Z Sun, Richard Socher, et~al.
\newblock Large language models generate functional protein sequences across diverse families.
\newblock \emph{Nature Biotechnology}, pages 1--8, 2023.

\bibitem[Mahbub and Bayzid(2022)]{mahbub2022egret}
Sazan Mahbub and Md~Shamsuzzoha Bayzid.
\newblock Egret: edge aggregated graph attention networks and transfer learning improve protein--protein interaction site prediction.
\newblock \emph{Briefings in Bioinformatics}, 23\penalty0 (2):\penalty0 bbab578, 2022.

\bibitem[Meier et~al.(2021)Meier, Rao, Verkuil, Liu, Sercu, and Rives]{ESM-1v}
Joshua Meier, Roshan Rao, Robert Verkuil, Jason Liu, Tom Sercu, and Alex Rives.
\newblock Language models enable zero-shot prediction of the effects of mutations on protein function.
\newblock \emph{Advances in Neural Information Processing Systems}, 34:\penalty0 29287--29303, 2021.

\bibitem[Micikevicius et~al.(2018)Micikevicius, Narang, Alben, Diamos, Elsen, Garcia, Ginsburg, Houston, Kuchaiev, Venkatesh, and Wu]{micikevicius2018mixed}
Paulius Micikevicius, Sharan Narang, Jonah Alben, Gregory Diamos, Erich Elsen, David Garcia, Boris Ginsburg, Michael Houston, Oleksii Kuchaiev, Ganesh Venkatesh, and Hao Wu.
\newblock Mixed precision training, 2018.

\bibitem[Ning et~al.(2023)Ning, Zhu, Xie, Lin, Cui, Yuan, Chen, and Yuan]{ning2023videobench}
Munan Ning, Bin Zhu, Yujia Xie, Bin Lin, Jiaxi Cui, Lu~Yuan, Dongdong Chen, and Li~Yuan.
\newblock Video-bench: A comprehensive benchmark and toolkit for evaluating video-based large language models, 2023.

\bibitem[Notin et~al.(2022)Notin, Dias, Frazer, Hurtado, Gomez, Marks, and Gal]{notin2022tranception}
Pascal Notin, Mafalda Dias, Jonathan Frazer, Javier~Marchena Hurtado, Aidan~N Gomez, Debora Marks, and Yarin Gal.
\newblock Tranception: protein fitness prediction with autoregressive transformers and inference-time retrieval.
\newblock In \emph{International Conference on Machine Learning}, pages 16990--17017, 2022.

\bibitem[OpenAI(2023)]{openai2023gpt4}
OpenAI.
\newblock Gpt-4 technical report.
\newblock \emph{Arxiv Preprint}, arXiv:2303.08774, 2023.
\newblock URL \url{https://arxiv.org/abs/2303.08774}.

\bibitem[OpenAI et~al.(2024)OpenAI, Achiam, Adler, Agarwal, Ahmad, Akkaya, Aleman, Almeida, Altenschmidt, Altman, Anadkat, Avila, Babuschkin, Balaji, Balcom, Baltescu, Bao, Bavarian, Belgum, Bello, Berdine, Bernadett-Shapiro, Berner, Bogdonoff, Boiko, Boyd, Brakman, Brockman, Brooks, Brundage, Button, Cai, Campbell, Cann, Carey, Carlson, Carmichael, Chan, Chang, Chantzis, Chen, Chen, Chen, Chen, Chen, Chess, Cho, Chu, Chung, Cummings, Currier, Dai, Decareaux, Degry, Deutsch, Deville, Dhar, Dohan, Dowling, Dunning, Ecoffet, Eleti, Eloundou, Farhi, Fedus, Felix, Fishman, Forte, Fulford, Gao, Georges, Gibson, Goel, Gogineni, Goh, Gontijo-Lopes, Gordon, Grafstein, Gray, Greene, Gross, Gu, Guo, Hallacy, Han, Harris, He, Heaton, Heidecke, Hesse, Hickey, Hickey, Hoeschele, Houghton, Hsu, Hu, Hu, Huizinga, Jain, Jain, Jang, Jiang, Jiang, Jin, Jin, Jomoto, Jonn, Jun, Kaftan, Łukasz Kaiser, Kamali, Kanitscheider, Keskar, Khan, Kilpatrick, Kim, Kim, Kim, Kirchner, Kiros, Knight, Kokotajlo, Łukasz Kondraciuk, Kondrich,
  Konstantinidis, Kosic, Krueger, Kuo, Lampe, Lan, Lee, Leike, Leung, Levy, Li, Lim, Lin, Lin, Litwin, Lopez, Lowe, Lue, Makanju, Malfacini, Manning, Markov, Markovski, Martin, Mayer, Mayne, McGrew, McKinney, McLeavey, McMillan, McNeil, Medina, Mehta, Menick, Metz, Mishchenko, Mishkin, Monaco, Morikawa, Mossing, Mu, Murati, Murk, Mély, Nair, Nakano, Nayak, Neelakantan, Ngo, Noh, Ouyang, O'Keefe, Pachocki, Paino, Palermo, Pantuliano, Parascandolo, Parish, Parparita, Passos, Pavlov, Peng, Perelman, de~Avila Belbute~Peres, Petrov, de~Oliveira~Pinto, Michael, Pokorny, Pokrass, Pong, Powell, Power, Power, Proehl, Puri, Radford, Rae, Ramesh, Raymond, Real, Rimbach, Ross, Rotsted, Roussez, Ryder, Saltarelli, Sanders, Santurkar, Sastry, Schmidt, Schnurr, Schulman, Selsam, Sheppard, Sherbakov, Shieh, Shoker, Shyam, Sidor, Sigler, Simens, Sitkin, Slama, Sohl, Sokolowsky, Song, Staudacher, Such, Summers, Sutskever, Tang, Tezak, Thompson, Tillet, Tootoonchian, Tseng, Tuggle, Turley, Tworek, Uribe, Vallone, Vijayvergiya,
  Voss, Wainwright, Wang, Wang, Wang, Ward, Wei, Weinmann, Welihinda, Welinder, Weng, Weng, Wiethoff, Willner, Winter, Wolrich, Wong, Workman, Wu, Wu, Wu, Xiao, Xu, Yoo, Yu, Yuan, Zaremba, Zellers, Zhang, Zhang, Zhao, Zheng, Zhuang, Zhuk, and Zoph]{openai2024gpt4}
OpenAI, Josh Achiam, Steven Adler, Sandhini Agarwal, Lama Ahmad, Ilge Akkaya, Florencia~Leoni Aleman, Diogo Almeida, Janko Altenschmidt, Sam Altman, Shyamal Anadkat, Red Avila, Igor Babuschkin, Suchir Balaji, Valerie Balcom, Paul Baltescu, Haiming Bao, Mohammad Bavarian, Jeff Belgum, Irwan Bello, Jake Berdine, Gabriel Bernadett-Shapiro, Christopher Berner, Lenny Bogdonoff, Oleg Boiko, Madelaine Boyd, Anna-Luisa Brakman, Greg Brockman, Tim Brooks, Miles Brundage, Kevin Button, Trevor Cai, Rosie Campbell, Andrew Cann, Brittany Carey, Chelsea Carlson, Rory Carmichael, Brooke Chan, Che Chang, Fotis Chantzis, Derek Chen, Sully Chen, Ruby Chen, Jason Chen, Mark Chen, Ben Chess, Chester Cho, Casey Chu, Hyung~Won Chung, Dave Cummings, Jeremiah Currier, Yunxing Dai, Cory Decareaux, Thomas Degry, Noah Deutsch, Damien Deville, Arka Dhar, David Dohan, Steve Dowling, Sheila Dunning, Adrien Ecoffet, Atty Eleti, Tyna Eloundou, David Farhi, Liam Fedus, Niko Felix, Simón~Posada Fishman, Juston Forte, Isabella Fulford, Leo
  Gao, Elie Georges, Christian Gibson, Vik Goel, Tarun Gogineni, Gabriel Goh, Rapha Gontijo-Lopes, Jonathan Gordon, Morgan Grafstein, Scott Gray, Ryan Greene, Joshua Gross, Shixiang~Shane Gu, Yufei Guo, Chris Hallacy, Jesse Han, Jeff Harris, Yuchen He, Mike Heaton, Johannes Heidecke, Chris Hesse, Alan Hickey, Wade Hickey, Peter Hoeschele, Brandon Houghton, Kenny Hsu, Shengli Hu, Xin Hu, Joost Huizinga, Shantanu Jain, Shawn Jain, Joanne Jang, Angela Jiang, Roger Jiang, Haozhun Jin, Denny Jin, Shino Jomoto, Billie Jonn, Heewoo Jun, Tomer Kaftan, Łukasz Kaiser, Ali Kamali, Ingmar Kanitscheider, Nitish~Shirish Keskar, Tabarak Khan, Logan Kilpatrick, Jong~Wook Kim, Christina Kim, Yongjik Kim, Jan~Hendrik Kirchner, Jamie Kiros, Matt Knight, Daniel Kokotajlo, Łukasz Kondraciuk, Andrew Kondrich, Aris Konstantinidis, Kyle Kosic, Gretchen Krueger, Vishal Kuo, Michael Lampe, Ikai Lan, Teddy Lee, Jan Leike, Jade Leung, Daniel Levy, Chak~Ming Li, Rachel Lim, Molly Lin, Stephanie Lin, Mateusz Litwin, Theresa Lopez, Ryan
  Lowe, Patricia Lue, Anna Makanju, Kim Malfacini, Sam Manning, Todor Markov, Yaniv Markovski, Bianca Martin, Katie Mayer, Andrew Mayne, Bob McGrew, Scott~Mayer McKinney, Christine McLeavey, Paul McMillan, Jake McNeil, David Medina, Aalok Mehta, Jacob Menick, Luke Metz, Andrey Mishchenko, Pamela Mishkin, Vinnie Monaco, Evan Morikawa, Daniel Mossing, Tong Mu, Mira Murati, Oleg Murk, David Mély, Ashvin Nair, Reiichiro Nakano, Rajeev Nayak, Arvind Neelakantan, Richard Ngo, Hyeonwoo Noh, Long Ouyang, Cullen O'Keefe, Jakub Pachocki, Alex Paino, Joe Palermo, Ashley Pantuliano, Giambattista Parascandolo, Joel Parish, Emy Parparita, Alex Passos, Mikhail Pavlov, Andrew Peng, Adam Perelman, Filipe de~Avila Belbute~Peres, Michael Petrov, Henrique~Ponde de~Oliveira~Pinto, Michael, Pokorny, Michelle Pokrass, Vitchyr~H. Pong, Tolly Powell, Alethea Power, Boris Power, Elizabeth Proehl, Raul Puri, Alec Radford, Jack Rae, Aditya Ramesh, Cameron Raymond, Francis Real, Kendra Rimbach, Carl Ross, Bob Rotsted, Henri Roussez,
  Nick Ryder, Mario Saltarelli, Ted Sanders, Shibani Santurkar, Girish Sastry, Heather Schmidt, David Schnurr, John Schulman, Daniel Selsam, Kyla Sheppard, Toki Sherbakov, Jessica Shieh, Sarah Shoker, Pranav Shyam, Szymon Sidor, Eric Sigler, Maddie Simens, Jordan Sitkin, Katarina Slama, Ian Sohl, Benjamin Sokolowsky, Yang Song, Natalie Staudacher, Felipe~Petroski Such, Natalie Summers, Ilya Sutskever, Jie Tang, Nikolas Tezak, Madeleine~B. Thompson, Phil Tillet, Amin Tootoonchian, Elizabeth Tseng, Preston Tuggle, Nick Turley, Jerry Tworek, Juan Felipe~Cerón Uribe, Andrea Vallone, Arun Vijayvergiya, Chelsea Voss, Carroll Wainwright, Justin~Jay Wang, Alvin Wang, Ben Wang, Jonathan Ward, Jason Wei, CJ~Weinmann, Akila Welihinda, Peter Welinder, Jiayi Weng, Lilian Weng, Matt Wiethoff, Dave Willner, Clemens Winter, Samuel Wolrich, Hannah Wong, Lauren Workman, Sherwin Wu, Jeff Wu, Michael Wu, Kai Xiao, Tao Xu, Sarah Yoo, Kevin Yu, Qiming Yuan, Wojciech Zaremba, Rowan Zellers, Chong Zhang, Marvin Zhang, Shengjia
  Zhao, Tianhao Zheng, Juntang Zhuang, William Zhuk, and Barret Zoph.
\newblock Gpt-4 technical report, 2024.

\bibitem[Organization and University(2007)]{world2007protein}
World~Health Organization and United~Nations University.
\newblock \emph{Protein and amino acid requirements in human nutrition}, volume 935.
\newblock World Health Organization, 2007.

\bibitem[Rao et~al.(2020)Rao, Meier, Sercu, Ovchinnikov, and Rives]{ESM}
Roshan Rao, Joshua Meier, Tom Sercu, Sergey Ovchinnikov, and Alexander Rives.
\newblock Transformer protein language models are unsupervised structure learners.
\newblock \emph{Biorxiv}, pages 2020--12, 2020.

\bibitem[R{'e}au et~al.(2023)R{'e}au, Renaud, Xue, and Bonvin]{reau2023deeprank}
Manon R{'e}au, Nicolas Renaud, Li~C Xue, and Alexandre~MJJ Bonvin.
\newblock Deeprank-gnn: a graph neural network framework to learn patterns in protein--protein interfaces.
\newblock \emph{Bioinformatics}, 39\penalty0 (1):\penalty0 btac759, 2023.

\bibitem[Rives et~al.(2021)Rives, Meier, Sercu, Goyal, Lin, Liu, Guo, Ott, Zitnick, Ma, et~al.]{ESM-1b}
Alexander Rives, Joshua Meier, Tom Sercu, Siddharth Goyal, Zeming Lin, Jason Liu, Demi Guo, Myle Ott, C~Lawrence Zitnick, Jerry Ma, et~al.
\newblock Biological structure and function emerge from scaling unsupervised learning to 250 million protein sequences.
\newblock \emph{Proceedings of the National Academy of Sciences}, 118\penalty0 (15):\penalty0 e2016239118, 2021.

\bibitem[Shu et~al.(2023)Shu, Zhang, Jiang, and Xie]{shu2023audiovisual}
Fangxun Shu, Lei Zhang, Hao Jiang, and Cihang Xie.
\newblock Audio-visual llm for video understanding, 2023.

\bibitem[Suzek et~al.(2015)Suzek, Wang, Huang, McGarvey, and Wu]{suzek2015uniref}
BE~Suzek, Y~Wang, H~Huang, PB~McGarvey, and CH~Wu.
\newblock Uniref clusters: a comprehensive and scalable alternative for improving sequence similarity searches.
\newblock \emph{Bioinformatics}, 2015.

\bibitem[Teague(2003)]{Teague2003}
Simon~J. Teague.
\newblock Implications of protein flexibility for drug discovery.
\newblock \emph{Nature Reviews Drug Discovery}, 2\penalty0 (7):\penalty0 527–541, July 2003.
\newblock ISSN 1474-1784.
\newblock \doi{10.1038/nrd1129}.
\newblock URL \url{http://dx.doi.org/10.1038/nrd1129}.

\bibitem[Team et~al.(2023)Team, Anil, Borgeaud, Wu, Alayrac, Yu, Soricut, Schalkwyk, Dai, Hauth, et~al.]{team2023gemini}
Gemini Team, Rohan Anil, Sebastian Borgeaud, Yonghui Wu, Jean-Baptiste Alayrac, Jiahui Yu, Radu Soricut, Johan Schalkwyk, Andrew~M Dai, Anja Hauth, et~al.
\newblock Gemini: a family of highly capable multimodal models.
\newblock \emph{arXiv:2312.11805}, 2023.

\bibitem[Touvron et~al.(2023)Touvron, Lavril, Izacard, Martinet, Lachaux, Lacroix, Rozière, Goyal, Hambro, Azhar, Rodriguez, Joulin, Grave, and Lample]{touvron2023llama}
Hugo Touvron, Thibaut Lavril, Gautier Izacard, Xavier Martinet, Marie-Anne Lachaux, Timothée Lacroix, Baptiste Rozière, Naman Goyal, Eric Hambro, Faisal Azhar, Aurelien Rodriguez, Armand Joulin, Edouard Grave, and Guillaume Lample.
\newblock Llama: Open and efficient foundation language models, 2023.

\bibitem[Vaswani et~al.(2017)Vaswani, Shazeer, Parmar, Uszkoreit, Jones, Gomez, Kaiser, and Polosukhin]{transformer}
Ashish Vaswani, Noam Shazeer, Niki Parmar, Jakob Uszkoreit, Llion Jones, Aidan~N Gomez, {\L}ukasz Kaiser, and Illia Polosukhin.
\newblock Attention is all you need.
\newblock \emph{Advances in neural information processing systems}, 30, 2017.

\bibitem[Vig et~al.(2020)Vig, Madani, Varshney, Xiong, Socher, and Rajani]{vig2020bertology}
Jesse Vig, Ali Madani, Lav~R Varshney, Caiming Xiong, Richard Socher, and Nazneen~Fatema Rajani.
\newblock Bertology meets biology: Interpreting attention in protein language models.
\newblock \emph{arXiv preprint arXiv:2006.15222}, 2020.

\bibitem[Wang et~al.(2019)Wang, Singh, Michael, Hill, Levy, and Bowman]{wang2019glue}
Alex Wang, Amanpreet Singh, Julian Michael, Felix Hill, Omer Levy, and Samuel~R. Bowman.
\newblock Glue: A multi-task benchmark and analysis platform for natural language understanding, 2019.

\bibitem[Wang et~al.(2024)Wang, Fan, Quan, and Yang]{wang2024protchatgpt}
Chao Wang, Hehe Fan, Ruijie Quan, and Yi~Yang.
\newblock Protchatgpt: Towards understanding proteins with large language models, 2024.

\bibitem[Xiao et~al.(2025)Xiao, Zhao, Zhang, Jin, Zhang, Ren, Sun, Wang, Wan, Lu, et~al.]{xiao2025protein}
Yijia Xiao, Wanjia Zhao, Junkai Zhang, Yiqiao Jin, Han Zhang, Zhicheng Ren, Renliang Sun, Haixin Wang, Guancheng Wan, Pan Lu, et~al.
\newblock Protein large language models: A comprehensive survey.
\newblock \emph{arXiv:2502.17504}, 2025.

\bibitem[Xu et~al.(2023)Xu, Yuan, Miret, and Tang]{xu2023protst}
Minghao Xu, Xinyu Yuan, Santiago Miret, and Jian Tang.
\newblock Protst: Multi-modality learning of protein sequences and biomedical texts, 2023.

\bibitem[Yao et~al.(2024)Yao, Jiang, Yang, and Hu]{yao2024benchmarking}
Binwei Yao, Ming Jiang, Diyi Yang, and Junjie Hu.
\newblock Benchmarking llm-based machine translation on cultural awareness, 2024.

\bibitem[Yu et~al.(2023)Yu, Yang, Li, Wang, Lin, Liu, Wang, and Wang]{yu2023mmvet}
Weihao Yu, Zhengyuan Yang, Linjie Li, Jianfeng Wang, Kevin Lin, Zicheng Liu, Xinchao Wang, and Lijuan Wang.
\newblock Mm-vet: Evaluating large multimodal models for integrated capabilities, 2023.

\bibitem[Zellers et~al.(2019)Zellers, Holtzman, Bisk, Farhadi, and Choi]{zellers2019hellaswag}
Rowan Zellers, Ari Holtzman, Yonatan Bisk, Ali Farhadi, and Yejin Choi.
\newblock Hellaswag: Can a machine really finish your sentence?, 2019.

\bibitem[Zhang et~al.(2022)Zhang, Bi, Liang, Cheng, Hong, Deng, Lian, Zhang, and Chen]{zhang2022ontoprotein}
Ningyu Zhang, Zhen Bi, Xiaozhuan Liang, Siyuan Cheng, Haosen Hong, Shumin Deng, Jiazhang Lian, Qiang Zhang, and Huajun Chen.
\newblock Ontoprotein: Protein pretraining with gene ontology embedding.
\newblock \emph{arXiv preprint arXiv:2201.11147}, 2022.

\bibitem[Zhang et~al.(2023{\natexlab{a}})Zhang, Han, Zhou, Hu, Yan, Lu, Li, Gao, and Qiao]{zhang2023llama}
Renrui Zhang, Jiaming Han, Aojun Zhou, Xiangfei Hu, Shilin Yan, Pan Lu, Hongsheng Li, Peng Gao, and Yu~Qiao.
\newblock Llama-adapter: Efficient fine-tuning of language models with zero-init attention.
\newblock \emph{arXiv preprint arXiv:2303.16199}, 2023{\natexlab{a}}.

\bibitem[Zhang et~al.(2020)Zhang, Kishore, Wu, Weinberger, and Artzi]{zhangbertscore}
Tianyi Zhang, Varsha Kishore, Felix Wu, Kilian~Q Weinberger, and Yoav Artzi.
\newblock Bertscore: Evaluating text generation with bert.
\newblock In \emph{ICLR}, 2020.

\bibitem[Zhang et~al.(2023{\natexlab{b}})Zhang, Ladhak, Durmus, Liang, McKeown, and Hashimoto]{zhang2023benchmarking}
Tianyi Zhang, Faisal Ladhak, Esin Durmus, Percy Liang, Kathleen McKeown, and Tatsunori~B. Hashimoto.
\newblock Benchmarking large language models for news summarization, 2023{\natexlab{b}}.

\bibitem[Zhang et~al.(2023{\natexlab{c}})Zhang, Xu, Jamasb, Chenthamarakshan, Lozano, Das, and Tang]{GearNet}
Zuobai Zhang, Minghao Xu, Arian Jamasb, Vijil Chenthamarakshan, Aurelie Lozano, Payel Das, and Jian Tang.
\newblock Protein representation learning by geometric structure pretraining.
\newblock In \emph{International Conference on Learning Representations}, 2023{\natexlab{c}}.

\end{thebibliography}

\appendix
\label{sec:appendix}

\section*{Limitations}
While \model demonstrates strong capabilities in protein sequence and structure understanding, there are areas that can be improved: 
1) \emph{Potential for Hallucination.} 
As with most LLM-based systems, \model may occasionally generate responses that are not fully aligned with established biological knowledge. However, its integration of protein-specific encoders helps mitigate this risk by grounding predictions in structured data. Future work can further refine this by incorporating confidence scores or uncertainty estimation mechanisms. 
% \model leverages the capabilities of LLMs for protein conversations, thus inheriting their potential for hallucination. This may imply safety concerns in the field of biology as it can provide faulty information with significant consequences. 
2) \emph{Verifiability.} Currently, \model does not provide direct citations for its responses, which may make it challenging to trace specific claims back to primary sources. 
We did not include citation data into the current training set of \model due to the scarcity of protein datasets with reliable, consistent, and fine-grained citations that link specific claims or answers to appropriate references. As a result, integrating this data could potentially lead to inaccurate answers.
Techniques such as retrieval augmented generation (RAG)~\citep{} or explicit literature grounding for better reliability and trustworthiness. 
3) \emph{Training Data}.
As with any data-driven model, 
the performance of \model is influenced by the quality and diversity of the training data. 
While we carefully curated a large-scale dataset of \datasize proteins with structured annotations, different processing \& alignment strategies, continuous enhancement, and feedback from biological experts can further optimize its performance. 

As demonstrated with alignment using GPT-4, different strategies may yield vastly different results.

\section{Open Access and Deployment}

To maximize accessibility, usability, and reproducibility, we will open-source both \model and the \dataset dataset, allowing researchers to experiment with different backbone LLMs and protein encoders. 
\model's flexible design ensures adaptability to model architectures.

\section{Additional Experiments}

\subsection{Case Studies}
\begin{figure*}[htbp]
% \vskip 0.2in
\begin{center}
\centerline{\includegraphics[width=0.85\textwidth]{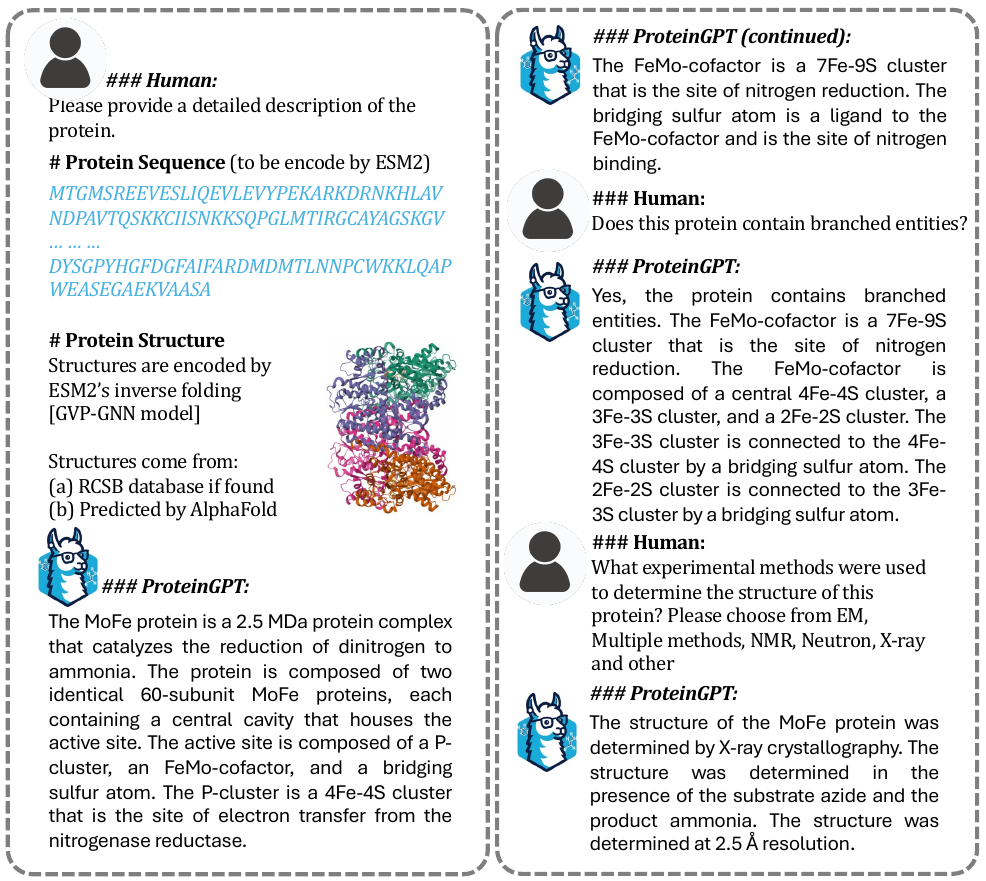}}
\caption{ProteinGPT's conversation on Protein 7RUV.}
\label{fig:7RUV}
\end{center}
% \vskip -0.2in
\end{figure*}

To avoid data leakage, we reserved 30\% of our QA and abstract dataset for testing, which is around 26,419 proteins. This ensures that the tests reflect real-world scenarios as \model has never seen these proteins before during training. We provide \model's Q and A on Protein 7RUV in Figure \ref{fig:7RUV}.

\section{Comparative Plots}
\begin{figure*}
    \centering
    \includegraphics[width=0.8\linewidth]{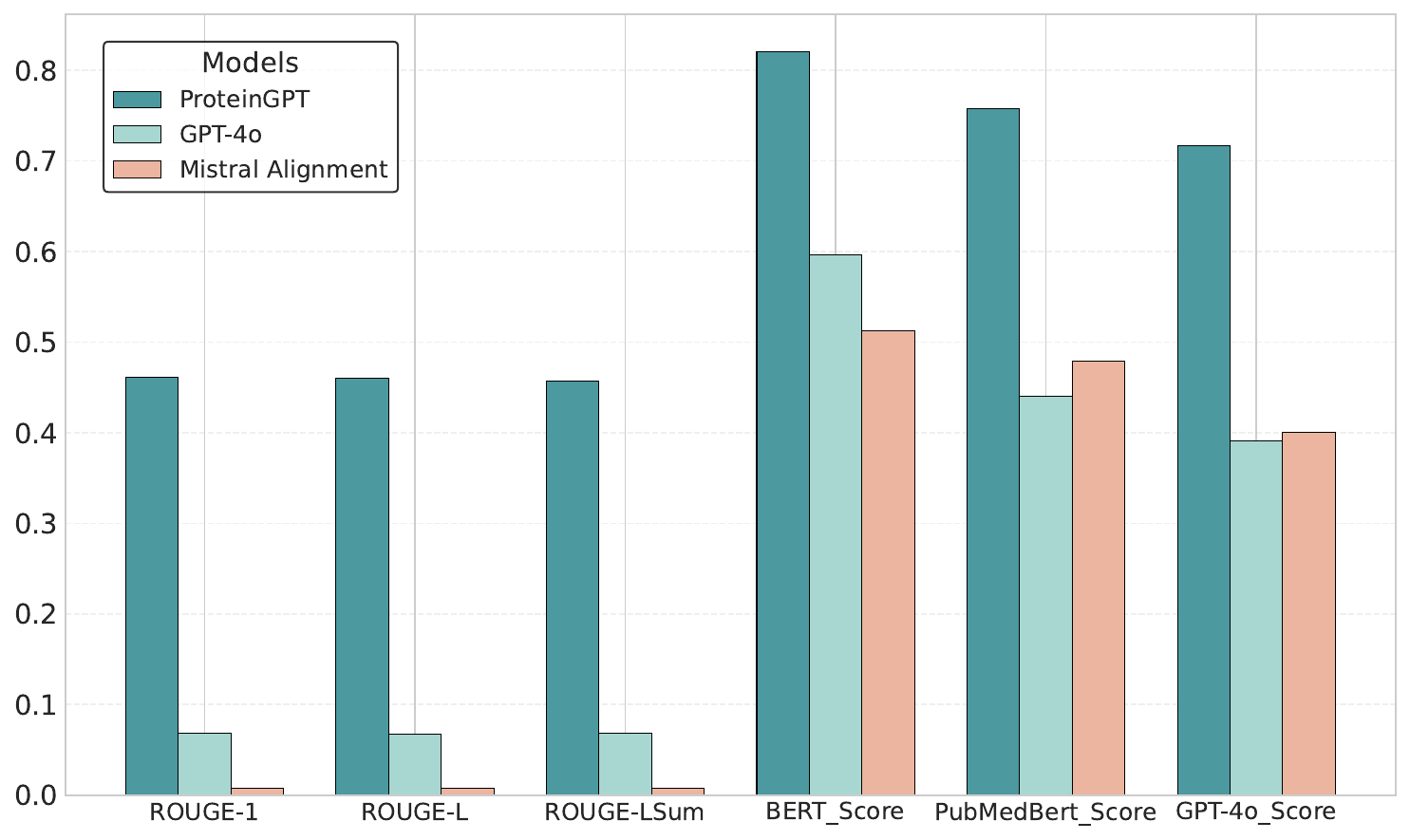}
    \caption{Comparison of \model, OpenAI's GPT, and Modality-Aligned-only model.}
    \label{fig:perform_compare}
\end{figure*}

Figures \ref{fig:vis_proteingpt} and \ref{fig:perform_compare} highlight the performance comparison across different models, including various versions of ProteinGPT, OpenAI's GPT models, and Mistral models, using multiple evaluation metrics.

\begin{figure*}[htbp]
% \vskip 0.2in
\begin{center}
\centerline{\includegraphics[width=\linewidth]{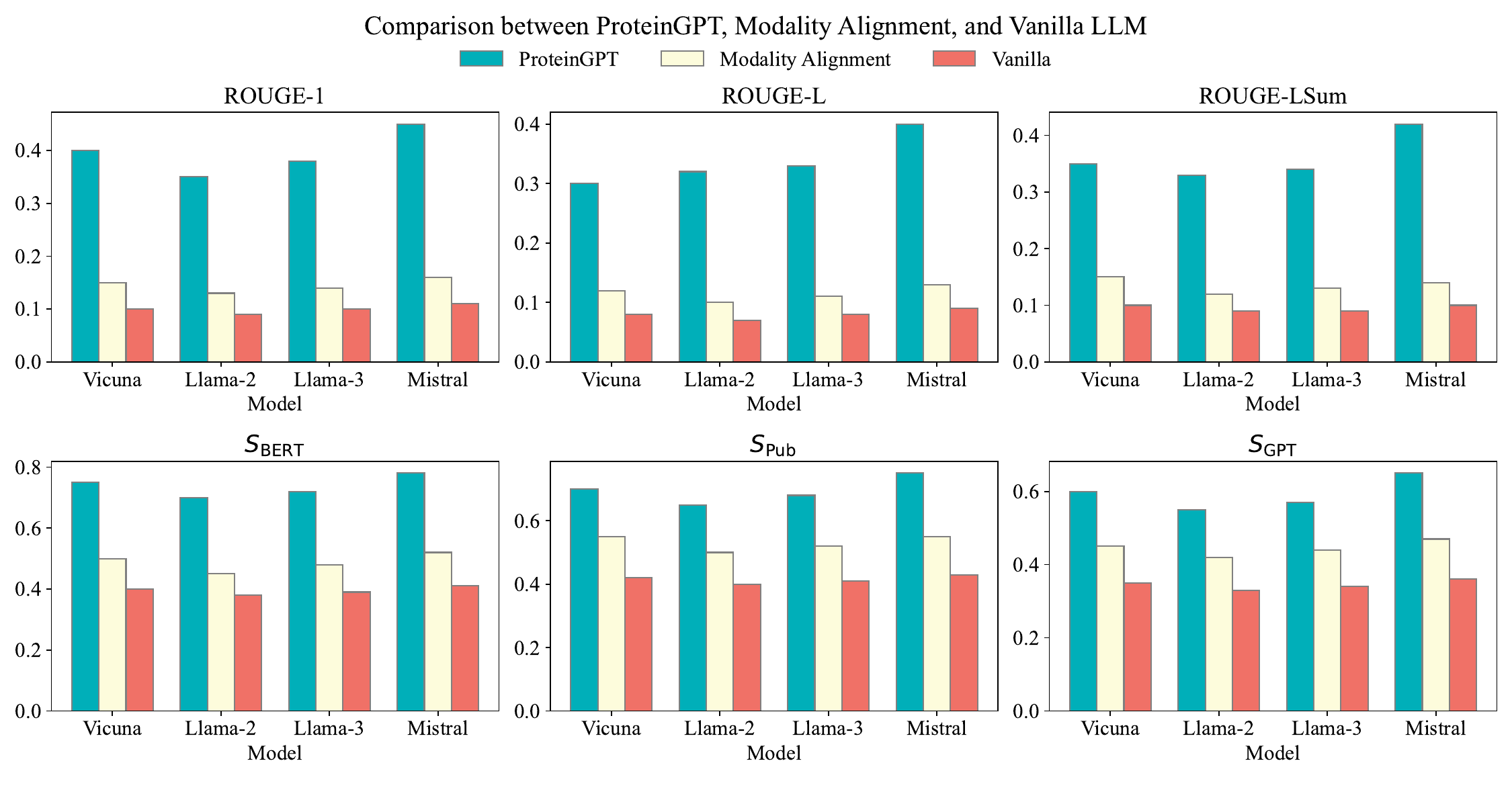}}
\caption{Comparison of Different Strategies and Models.}
\label{fig:perform_it}
\end{center}
% \vskip -0.2in
\end{figure*}

Figure \ref{fig:perform_compare} shows the performance of ProteinGPT variants alongside OpenAI's GPT and MA models. The key observations from this figure are:

\begin{itemize}[leftmargin=1em]
    \item \textbf{ROUGE-1 and ROUGE-L}: The ProteinGPT\_Mistral model significantly outperforms the other models with scores of 0.451 and 0.460, respectively. ProteinGPT\_LLaMA-3 and the other GPT variants (GPT-3.5-turbo, GPT-4-turbo, GPT-4o) have much lower scores, indicating that the Mistral-based variant is superior in these metrics.
    \item \textbf{ROUGE-LSum}: Similar to ROUGE-1 and ROUGE-L, the ProteinGPT\_Mistral variant leads with a score of 0.457, followed closely by ProteinGPT\_LLaMA-3 at 0.387. Other models show significantly lower scores, emphasizing the effectiveness of the Mistral variant.
    \item \textbf{BERT Score} ($S_{\mathrm{BERT}}$): The ProteinGPT\_Mistral model also performs best with a score of 0.821, with ProteinGPT\_LLaMA-3 following at 0.779. The GPT models lag behind, demonstrating that the ProteinGPT variants are more aligned with human evaluations.
    \item \textbf{PubMedBert Score} ($S_{\mathrm{Pub}}$): Again, ProteinGPT\_Mistral achieves the highest score of 0.758, outperforming ProteinGPT\_LLaMA-3 slightly. The GPT models perform lower in this biomedical domain-specific metric.
    \item \textbf{GPT-4o Score}  ($S_{\mathrm{GPT}}$): The trend continues with ProteinGPT\_Mistral leading at 0.717, while ProteinGPT\_LLaMA-3 scores 0.706, suggesting a close performance in this category.
\end{itemize}

In Figure \ref{fig:vis_proteingpt}, the performance of ProteinGPT with different base large language models (LLMs) is evaluated. The following findings are noteworthy:

\begin{itemize}[leftmargin=1em]
    \item \textbf{ROUGE-1 and ROUGE-L}: The Mistral variant of ProteinGPT outperforms other base models (Vicuna, LLaMA-2, and LLaMA-3) in both ROUGE-1 (0.461) and ROUGE-L (0.460) scores. LLaMA-3 follows, but with slightly lower performance, while Vicuna and LLaMA-2 have the lowest scores.
    \item \textbf{ROUGE-LSum}: The Mistral base again outperforms others with a score of 0.457, while the LLaMA-3 variant is close behind at 0.367.
    \item \textbf{BERT Score}: Mistral shows the best performance (0.821), with other bases following in descending order: LLaMA-3 (0.779), LLaMA-2 (0.730), and Vicuna (0.739).
    \item \textbf{PubMedBert Score}: Mistral continues to outperform other base models with a score of 0.758, while LLaMA-3 follows at 0.706.
    \item \textbf{GPT-4o Score}: The Mistral base model scores highest at 0.717, followed by LLaMA-3 at 0.698.
\end{itemize}

Overall, the findings suggest that the ProteinGPT\_Mistral variant consistently outperforms other models across all metrics. The Mistral base model proves to be the most effective for ProteinGPT, indicating that the combination of Mistral with ProteinGPT architecture is particularly powerful. In contrast, other base models such as LLaMA-2 and Vicuna, and the GPT variants, show comparatively lower performance, especially in metrics like ROUGE-1, ROUGE-L, and the PubMedBert Score, which are crucial for natural language processing tasks in specialized domains such as biomedical text analysis.

% \section{Comparative Plots}

% \begin{figure*}[htbp]
% % \vskip 0.2in
% \begin{center}
% \centerline{\includegraphics[width=0.7\linewidth]{figures/ProteinGPT_BaseLLM_Comparison_minimal.pdf}}
% \caption{ProteinGPT's Performance with Different Base LLMs. }
% \label{fig:vis_proteingpt}
% \end{center}
% % \vskip -0.2in
% \end{figure*}
% \begin{figure*}[htbp]
% % \vskip 0.2in
% \begin{center}
% \centering
% \includegraphics[width=0.7\linewidth]{figures/ProteinGPT_Performance_Comparison_minimal.pdf}
% \caption{Comparison of ProteinGPT, OpenAI's GPT and Modality-Aligned-only model.}
% % \caption{Comparison of ProteinGPT, OpenAI's GPT and MA models\footnote{LLM with modality alignment, but without instruction tuning.}.}
% \label{fig:perform_compare}
% \end{center}
% % \vskip -0.2in
% \end{figure*}

% \begin{figure*}[htbp]
% % \vskip 0.2in
% \begin{center}
% \centerline{\includegraphics[width=\linewidth]{figures/Comparison.pdf}}
% \caption{Comparison of Different Strategies and Models.}
% \label{fig:perform_it}
% \end{center}
% % \vskip -0.2in
% \end{figure*}

\begin{figure*}[htbp]
% \vskip 0.2in
\begin{center}
\centerline{\includegraphics[width=0.9\textwidth]{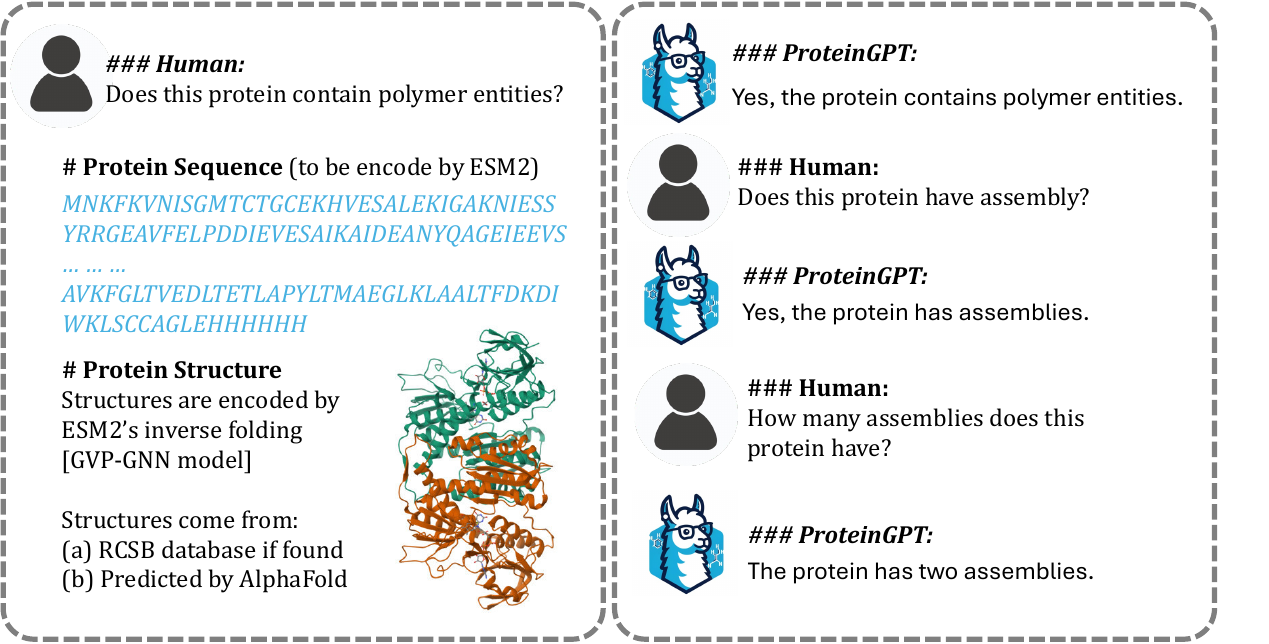}}
\caption{\model conversation on Protein 5X1Y.}
\label{fig:5X1Y}
\end{center}
% \vskip -0.2in
\end{figure*}

% More analysis in Section \ref{sec:appendix}.

\ifthenelse{\boolean{forKDD}}{
}{

}

% \begin{figure}[ht]
% \vskip 0.2in
% \begin{center}
%     \begin{subfigure}{0.5\columnwidth}
%         \centering
%         \includegraphics[width=\columnwidth]{figures/conversation1.pdf}
%         \caption{Conversation 1}
%         \label{fig:conv1}
%     \end{subfigure}%
%     \begin{subfigure}{0.5\columnwidth}
%         \centering
%         \includegraphics[width=\columnwidth]{figures/conversation1.pdf}
%         \caption{Conversation 2}
%         \label{fig:conv2}
%     \end{subfigure}
% \caption{Model Conversations}
% \label{fig:conversations}
% \end{center}
% \vskip -0.2in
% \end{figure}

\definecolor{lightblue}{rgb}{0.85, 0.92, 1.0}   % soft light blue, for background or highlights
\definecolor{lightgray}{rgb}{0.95, 0.95, 0.95}   % very light gray, for subtle contrasts
\definecolor{lightgreen}{rgb}{0.80, 0.92, 0.80}  % soft light green, for positive highlights
\definecolor{lightyellow}{rgb}{1.0, 1.0, 0.85}   % pale light yellow, for attention-catching highlights
\definecolor{lightpink}{rgb}{1.0, 0.85, 0.90}    % soft pink, for gentle emphasis
\definecolor{lightpurple}{rgb}{0.75, 0.65, 1.0}  % muted light purple, for distinguishing sections
\definecolor{lightorange}{rgb}{1.0, 0.90, 0.80}  % light orange, for calls to action or warnings
\definecolor{lightred}{rgb}{1.0, 0.80, 0.80}     % light red, for critical notes or errors
\definecolor{brightblue}{rgb}{0.45, 0.65, 1.0}   % medium bright blue, for links or links to references
\definecolor{lightlilac}{rgb}{0.85, 0.75, 1.0}  % soft lilac, for decorative or neutral accents

\lstdefinestyle{agent1}{
    backgroundcolor=\color{lightblue}, 
    basicstyle=\ttfamily\footnotesize,
    frame=single,
    rulecolor=\color{black},
    xleftmargin=0.5cm,
    framexleftmargin=0.5cm,
    numbers=left,
    numberstyle=\tiny\color{black},
    stepnumber=1,
    numbersep=5pt,
    lineskip=1pt
}

\lstdefinestyle{agent2}{
    backgroundcolor=\color{lightgray}, 
    basicstyle=\ttfamily\footnotesize,
    frame=single,
    rulecolor=\color{black},
    xleftmargin=0.5cm,
    framexleftmargin=0.5cm,
    numbers=left,
    numberstyle=\tiny\color{black},
    stepnumber=1,
    numbersep=5pt,
    lineskip=1pt
}

\lstdefinestyle{agent3}{
    backgroundcolor=\color{lightgreen}, 
    basicstyle=\ttfamily\footnotesize,
    frame=single,
    rulecolor=\color{black},
    xleftmargin=0.5cm,
    framexleftmargin=0.5cm,
    numbers=left,
    numberstyle=\tiny\color{black},
    stepnumber=1,
    numbersep=5pt,
    lineskip=1pt
}

% \newpage
\onecolumn

\definecolor{lightblue}{rgb}{0.85, 0.92, 1.0}   % soft light blue, for background or highlights
\definecolor{lightgray}{rgb}{0.95, 0.95, 0.95}   % very light gray, for subtle contrasts
\definecolor{lightgreen}{rgb}{0.80, 0.92, 0.80}  % soft light green, for positive highlights
\definecolor{lightyellow}{rgb}{1.0, 1.0, 0.85}   % pale light yellow, for attention-catching highlights
\definecolor{lightpink}{rgb}{1.0, 0.85, 0.90}    % soft pink, for gentle emphasis
\definecolor{lightpurple}{rgb}{0.75, 0.65, 1.0}  % muted light purple, for distinguishing sections
\definecolor{lightorange}{rgb}{1.0, 0.90, 0.80}  % light orange, for calls to action or warnings
\definecolor{lightred}{rgb}{1.0, 0.80, 0.80}     % light red, for critical notes or errors
\definecolor{brightblue}{rgb}{0.45, 0.65, 1.0}   % medium bright blue, for links or links to references
\definecolor{lightlilac}{rgb}{0.85, 0.75, 1.0}  % soft lilac, for decorative or neutral accents

\lstdefinestyle{agent1}{
    backgroundcolor=\color{lightblue}, 
    basicstyle=\ttfamily\footnotesize,
    frame=single,
    rulecolor=\color{black},
    xleftmargin=0.5cm,
    framexleftmargin=0.5cm,
    numbers=left,
    numberstyle=\tiny\color{black},
    stepnumber=1,
    numbersep=5pt,
    lineskip=1pt
}

\lstdefinestyle{agent2}{
    backgroundcolor=\color{lightgray}, 
    basicstyle=\ttfamily\footnotesize,
    frame=single,
    rulecolor=\color{black},
    xleftmargin=0.5cm,
    framexleftmargin=0.5cm,
    numbers=left,
    numberstyle=\tiny\color{black},
    stepnumber=1,
    numbersep=5pt,
    lineskip=1pt
}

\lstdefinestyle{agent3}{
    backgroundcolor=\color{lightgreen}, 
    basicstyle=\ttfamily\footnotesize,
    frame=single,
    rulecolor=\color{black},
    xleftmargin=0.5cm,
    framexleftmargin=0.5cm,
    numbers=left,
    numberstyle=\tiny\color{black},
    stepnumber=1,
    numbersep=5pt,
    lineskip=1pt
}

\onecolumn

\section{Case Studies on OpenAI and DeepSeek models}

\subsection{OpenAI o1-mini Model}
\label{app:o1_mini_answers}
\vspace{2mm}

Below are OpenAI's o1-mini responses to questions on protein 6RE5's property and structure.

\vspace{2mm}

\textbf{I. System Prompt}

<SYSTEM> You are an expert in protein domain, please answer the questions given the protein primary sequence and only give a concise answer to the question. </SYSTEM>
\vspace{2mm}

\textbf{II. Protein sequence information}

\texttt{>6RE5\_1|Chain A[auth 0]|ASA-10: Polytomella F-ATP synthase associated subunit 10|Polytomella sp. Pringsheim 198.80 (37502)}

\seqsplit{MSYSAYFAKAGFQFPAGLSALVAGIVALNVCTGRPTKGTKEISNAEYNATPIGYLQSPDQHPTAFPKVPGMKDVHGSPHHHH}

\vspace{2mm}

\textbf{III. Question Group 1}

\vspace{2mm}

\begin{lstlisting}[style=agent1, breaklines=true]
> Question: How many assemblies does this protein have?
> Answer (o1-mini): The assembly state of the protein cannot be determined solely from the primary amino acid sequence provided.

> Question: How many entities does this protein have?
> Answer (o1-mini): The protein has **two entities**: the main protein sequence and the His-tag (HHHH).

> Question: How many heavy atom coordinates records does this protein have?
> Answer (o1-mini): The protein has **562** heavy atom coordinates records.

> Question: How many hydrogen atom coordinates records does this protein have?
> Answer (o1-mini): Hydrogen atom coordinates are not available from the primary amino acid sequence alone.

> Question: How many model structures deposited for this protein?
> Answer (o1-mini): As of October 2023, there are no model structures deposited for this protein sequence.

> Question: How many polymer monomers does this protein have?
> Answer (o1-mini): The protein consists of **82 amino acid monomers**.

> Question: How many heavy solvent atom coordinates records does this protein have?
> Answer (o1-mini): The number of heavy solvent atom coordinates records cannot be determined from the primary sequence alone.

> Question: How many nucleic acid polymer entities (DNA or RNA) does this protein have?
> Answer (o1-mini): The protein does not contain any nucleic acid polymer entities (DNA or RNA).

> Question: How many intermolecular covalent bonds does this protein have?
> Answer (o1-mini): This protein has **zero intermolecular covalent bonds**.

> Question: How many intermolecular metalic bonds does this protein have?
> Answer (o1-mini): Based on the provided sequence, the protein contains four histidine residues (`HHHH`) that are potential sites for intermolecular metal coordination bonds. Therefore, it can form up to **four intermolecular metal bonds**.

> Question: Does this protein contain non-polymer entities?
> Answer (o1-mini): No, based on the provided primary sequence, there are no indications of non-polymer entities.

> Question: Does this protein contain polymer entities?
> Answer (o1-mini): No, the protein does not contain polymer entities.

> Question: Does this protein contain DNA polymer entities?
> Answer (o1-mini): No, the provided protein sequence does not contain DNA polymerase domains.

> Question: Does this protein contain RNA polymer entities?
> Answer (o1-mini): No, this protein does not contain RNA polymerase domains.

> Question: Does this protein contain solvent entities?
> Answer (o1-mini): Yes, the protein contains residues (such as polar and charged amino acids) that are likely to interact with the solvent.

> Question: Does this protein contain branched entities?
> Answer (o1-mini): Yes, the protein contains several branched amino acids, including valine (V), isoleucine (I), and leucine (L).

> Question: Does this protein have unmodeled polymer monomers?
> Answer (o1-mini): Yes, the protein contains a His-tag sequence (HHHH) at the C-terminus, which may represent unmodeled polymer monomers.

> Question: Does this protein have hybrid nucleic acid polymer entities?
> Answer (o1-mini): No, based on the provided primary amino acid sequence, there is no indication of hybrid nucleic acid polymer entities.

> Question: Does this protein have cis-peptide linkages?
> Answer (o1-mini): Based solely on the primary sequence, the presence of proline residues (P) suggests potential sites for cis-peptide linkages. However, without structural data, it's not possible to definitively confirm the existence of cis-peptide bonds in this protein.
\end{lstlisting}

\vspace{2.5mm}

\textbf{III. Question Group 2}

\vspace{2.5mm}
\begin{lstlisting}[style=agent2, breaklines=true]

> Question: What is the polymer entity composition for this protein?
> Answer (o1-mini): The protein consists of the following amino acids:

- **Alanine (A):** 10
- **Arginine (R):** 1
- **Asparagine (N):** 3
- **Aspartic acid (D):** 2
- **Cysteine (C):** 1
- **Glutamine (Q):** 3
- **Glutamic acid (E):** 2
- **Glycine (G):** 8
- **Histidine (H):** 6
- **Isoleucine (I):** 3
- **Leucine (L):** 4
- **Lysine (K):** 5
- **Methionine (M):** 2
- **Phenylalanine (F):** 4
- **Proline (P):** 8
- **Serine (S):** 6
- **Threonine (T):** 5
- **Tyrosine (Y):** 4
- **Valine (V):** 5

> Question: What is the nucleic acid polymer entity type for this protein?
> Answer (o1-mini): The nucleic acid polymer entity type for this protein is **messenger RNA (mRNA)**.

> Question: What are the bound nonpolymer components for this protein
> Answer (o1-mini): The protein likely binds nickel (\text{Ni}^{2+}) or cobalt (\text{Co}^{2+}) ions through its polyhistidine (HHHH) tag.

> Question: What experimental method(s) were used to determine the structure of this protein?
> Answer (o1-mini): Based solely on the provided primary amino acid sequence, the specific experimental method(s) used to determine the protein's structure cannot be identified. Additional information, such as structural database entries or publication references, is required to ascertain the techniques employed.

> Question: What is the molecular mass (KDa) of polymer and non-polymer entities (exclusive of solvent) for this protein?
> Answer (o1-mini): The molecular mass of the protein polymer is approximately **8.82 kDa**. There are no significant non-polymer entities associated with this protein.

> Question: What is the polymer entity type for this protein?
> Answer (o1-mini): **Polymer Entity Type:** Polypeptide

> Question: What are the software programs reported in connection with the production of this protein?
> Answer (o1-mini): There are no specific software programs reported in connection with the production of this protein based solely on the provided primary sequence.

> Question: Is this protein determined by experimental or computational methods?
> Answer (o1-mini): It cannot be determined solely from the primary sequence.

> Question: What are the terms characterizing the protein?
> Answer (o1-mini): - **Polyhistidine Tag**: Contains a His\_4 tag at the C-terminus (HHHH).
- **Molecular Weight**: Approximately 10.7 kDa.
- **Isoelectric Point (pI)**: Around 6.2.
- **Amino Acid Composition**: Rich in alanine, glycine, and proline.
- **Secondary Structure**: Likely includes alpha helices and coiled regions.
- **Potential Functional Motifs**: May contain sites for phosphorylation or binding.
- **Solubility**: Predicted to be moderately soluble.
- **Expression Tag**: Suitable for affinity purification using nickel columns.

> Question: When is this protein first published?
> Answer (o1-mini): The first publication date of this protein cannot be determined solely from its primary amino acid sequence. Additional information such as the protein name, gene identifier, or related literature is required.
\end{lstlisting}

\subsection{OpenAI o3-mini Model}
\label{app:o3_mini_answers}
\vspace{2mm}

Below are OpenAI's o3-mini responses to questions on protein 1GC9's property and structure.

\vspace{2mm}

\textbf{I. System Prompt}

<SYSTEM> You are an expert in protein domain, please answer the questions given the protein primary sequence and only give a concise answer to the question. </SYSTEM>
\vspace{2mm}

\textbf{II. Protein sequence information}

\texttt{>1GC9\_1|Chain A|3-ISOPROPYLMALATE DEHYDROGENASE|Thermus thermophilus (300852)}

\seqsplit{MKVAVLPGDGIGPEVTEAALKVLRALDEAEGLGLAYEVFPFGGAAIDAFGEPFPEPTRKGVEEAEAVLLGSVGGPKWDGLPRKIRPETGLLSLRKSQDLFANLRPAKVFPGLERLSPLKEEIARGVDVLIVRELTGGIYFGEPRGMSEAEAWNTERYSKPEVERVARVAFEGARKRRKHVVSVDKANVLEVGEFWRKTVEEVGRGYPDVALEHQYVDAMAMHLVRSPARFDVVVTGNIFGDILSDLASVLPGSLGLLPSASLGRGTPVFEPVHGSAPDIAGKGIANPTAAILSAAMMLEHAFGLVELARKVEDAVAKALLETPPPDLGGSAGTEAFTATVLRHLA}

\vspace{2mm}

\textbf{III. Question Group 1}

\vspace{2mm}

\begin{lstlisting}[style=agent1, breaklines=true]
> Question: How many assemblies does this protein have?
> Answer (o3-mini): It forms a single, monomeric assembly.

> Question: How many entities does this protein have?
> Answer (o3-mini): This protein is a single-chain entity.

> Question: How many heavy atom coordinates records does this protein have?
> Answer (o3-mini): There are 0 heavy atom coordinate records, as only the primary sequence is provided.

> Question: How many hydrogen atom coordinates records does this protein have?
> Answer (o3-mini): This protein's primary sequence does not include any hydrogen atom coordinate records, so the answer is 0.

> Question: How many model structures deposited for this protein?
> Answer (o3-mini): There are no deposited model structures for this protein.

> Question: How many polymer monomers does this protein have?
> Answer (o3-mini): The protein consists of 349 amino acid residues.

> Question: How many heavy solvent atom coordinates records does this protein have?
> Answer (o3-mini): 0 heavy solvent atom coordinates records.

> Question: How many nucleic acid polymer entities (DNA or RNA) does this protein have?
> Answer (o3-mini): This protein contains no nucleic acid polymer entities.

> Question: How many intermolecular covalent bonds does this protein have?
> Answer (o3-mini): This protein has no intermolecular covalent bonds.

> Question: How many intermolecular metalic bonds does this protein have?
> Answer (o3-mini): 0

There is no evidence from the primary sequence indicating the formation of any intermolecular metallic bonds.

> Question: Does this protein contain non-polymer entities?
> Answer (o3-mini): No. The sequence only represents a polypeptide chain without any non-polymer entities.

> Question: Does this protein contain polymer entities?
> Answer (o3-mini): No, the protein is a single polypeptide chain composed solely of amino acids and does not include additional polymer entities.

> Question: Does this protein contain DNA polymer entities?
> Answer (o3-mini): No, this protein does not contain DNA polymer domains.

> Question: Does this protein contain RNA polymer entities?
> Answer (o3-mini): No, this protein does not seem to contain RNA polymerase domains.
\end{lstlisting}

\vspace{2.5mm}

\textbf{III. Question Group 2}

\vspace{2.5mm}
\begin{lstlisting}[style=agent2, breaklines=true]

> Question: Does this protein contain solvent entities?
> Answer (o3-mini): No. The sequence lists only the amino acids without any solvent molecules or other non-protein entities.

> Question: Does this protein contain branched entities?
> Answer (o3-mini): No, this protein is a single linear polypeptide and does not contain branched entities.

> Question: Does this protein have unmodeled polymer monomers?
> Answer (o3-mini): No, the sequence appears complete with all expected polymer residues modeled.

> Question: Does this protein have hybrid nucleic acid polymer entities?
> Answer (o3-mini): No, this sequence represents a protein, not a hybrid nucleic acid polymer entity.

> Question: Does this protein have cis-peptide linkages?
> Answer (o3-mini): Based solely on the primary sequence, there's no definitive evidence for cis-peptide bonds. Cis conformations - most often seen in X-Pro linkages-require structural data to confirm their presence.

> Question: What is the polymer entity composition for this protein?
> Answer (o3-mini): The protein is a single polypeptide chain composed of 346 standard L-amino acid residues.

> Question: What is the nucleic acid polymer entity type for this protein?
> Answer (o3-mini): This molecule is not a nucleic acid - it is a protein (an amino acid polymer).

> Question: What experimental method(s) were used to determine the structure of this protein?
> Answer (o3-mini): No experimental structure determination data is provided; only the primary sequence is given, so no method (e.g., X-ray crystallography or NMR spectroscopy) can be inferred.

> Question: What is the molecular mass (KDa) of polymer and non-polymer entities (exclusive of solvent) for this protein?
> Answer (o3-mini): The 344-residue chain weighs roughly 38 kDa, and there are no non-polymer entities present.

> Question: What is the polymer entity type for this protein?
> Answer (o3-mini): It is a protein, which is a polypeptide chain.

> Question: What are the software programs reported in connection with the production of this protein?
> Answer (o3-mini): No software programs were reported in connection with the production of this protein.

> Question: Is this protein determined by experimental or computational methods?
> Answer (o3-mini): There's no accompanying annotation indicating experimental validation, so based solely on the primary sequence it's unclear whether it was determined experimentally or computationally.

> Question: When is this protein first published?
> Answer (o3-mini): The provided sequence does not include publication details, so the first publication date cannot be determined without additional reference data.
    
\end{lstlisting}

\subsection{DeepSeek R1 Model}
\label{app:deepseek_answers}
\vspace{2mm}

Below are DeepSeek's R1 model responses to questions on protein 5XSP's property and structure.

\vspace{2mm}

\textbf{I. System Prompt}

<SYSTEM> You are an expert in protein domain, please answer the questions given the protein primary sequence and only give a concise answer to the question. </SYSTEM>
\vspace{2mm}

\textbf{II. Protein sequence information}

\texttt{>5XSP\_1|Chains A, B|Phosphodiesterase acting on cyclic dinucleotides|Staphylococcus aureus (1280)}

\seqsplit{GSMRTRVRARVISHALKDILAEGDKVIIMGHKRPDLDAIGAAIGVSRFAMMNNLEAYIVLNETDIDPTLRRVMNEIDKKPELRERFITSDDAWDMMTSKTTVVIVDTHKPELVLDENVLNKANRKVVIDHHRRGESFISNPLLIYMEPYASSTAELVTELLEYQPTEQRLTRLESTVMYAGIIVDTRNFTLRTGSRTFDAASYLRAHGADTILTQHFLKDDVDTYINRSELIRTVKVEDNGIAIAHGSDDKIYHPVTVAQAADELLSLEGIEASYVVARREDNLIGISARSLGSVNVQLTMEALGGGGHLTNAATQLKGVTVEEAIAQLQQAITEQLSRSEDA}

\vspace{2mm}

\textbf{III. Question 1: Does this protein have cis-peptide linkages?}

\vspace{2mm}

% \vspace{2.5mm}
\begin{lstlisting}[style=agent1, breaklines=true]
<think>
Okay, so I need to figure out whether the given protein has any cis-peptide linkages. I remember that in proteins, peptide bonds can be either cis or trans. The trans configuration is the most common because it's more energetically favorable. Cis peptide bonds are less common but can occur under certain conditions, often when specific residues are involved.

Looking at the primary sequence provided, I should check for any residues that are known to favor cis peptide bonds. The key residues I remember are proline (Pro, P) and glycine (Gly, G). Proline is particularly known for its ability to adopt the cis configuration because of its ring structure, which restricts its conformational flexibility. Glycine, being the smallest amino acid with a lot of flexibility, can also sometimes be found in cis peptide bonds, though less commonly than proline.

I'll scan through the sequence to spot any Pro or Gly residues. Let me go through each part:

- The sequence starts with:

GSMRTRVRARVISHALKDILAEGDKVIIMGHKRPDLDAIGAAIGVSRFAMMNNLEAYIVLNETDIDPTLRRVMNEIDKKPELRERFITSDDAWDMMTSKTTVVIVDTHKPELVL
DENVLNKANRKVVIDHHRRGESFISNPLLIYMEPYASSTAELVTELLEYQPTEQRLTRLESTVMYAGIIVDTRNFTLRTGSRTFDAASYLRAHGADTILTQHFLKDDVDTYINR
SELIRTVKVEDNGIAIAHGSDDKIYHPVTVAQAADELLSLEGIEASYVVARREDNLIGISARSLGSVNVQLTMEALGGGGHLTNAATQLKGVTVEEAIAQLQQAITEQLSRSEDA

Looking for Pro (P):

- I see a Pro at position ... let me count. Starting from the beginning: G, S, M, R, T, R, V, R, A, R, V, I, S, H, A, L, K, D, I, L, A, E, G, D, K, V, I, I, M, G, H, K, R, P, D, L, D, A, I, G, A, A, I, G, V, S, R, F, A, M, M, N, N, L, E, A, Y, I, V, L, N, E, T, D, I, D, P, T, L, R, R, V, M, N, E, I, D, K, K, P, E, L, R, E, R, F, I, T, S, D, D, A, W, D, M, M, T, S, K, T, T, V, V, I, V, D, T, H, K, P, E, L, V, L, D, E, N, V, L, N, K, A, N, R, K, V, V, I, D, H, H, R, R, G, E, S, F, I, S, N, P, L, L, I, Y, M, E, P, Y, A, S, S, T, A, E, L, V, T, E, L, L, E, Y, Q, P, T, E, Q, R, L, T, R, L, E, S, T, V, M, Y, A, G, I, I, V, D, T, R, N, F, T, L, R, T, G, S, R, T, F, D, A, A, S, Y, L, R, A, H, G, A, D, T, I, L, T, Q, H, F, L, K, D, D, V, D, T, Y, I, N, R, S, E, L, I, R, T, V, K, V, E, D, N, G, I, A, I, A, H, G, S, D, D, K, I, Y, H, P, V, T, V, A, Q, A, A, D, E, L, L, S, L, E, G, I, E, A, S, Y, V, V, A, R, R, E, D, N, L, I, G, I, S, A, R, S, L, G, S, V, N, V, Q, L, T, M, E, A, L, G, G, G, G, H, L, T, N, A, A, T, Q, L, K, G, V, T, V, E, E, A, I, A, Q, L, Q, Q, A, I, T, E, Q, L, S, R, S, E, D, A.

So, I see several Pro residues. Let me note their positions:

1. The first Pro is at position 34 (counting from 0 or 1? Let me check: the sequence is 0-based or 1-based. Well, in the sequence given, the 34th character is P. Let's assume it's the 35th residue if starting count from 1.

Wait, actually, in the sequence provided, the 34th character is P. Let me count from the start:

G(1), S(2), M(3), R(4), T(5), R(6), V(7), R(8), A(9), R(10), V(11), I(12), S(13), H(14), A(15), L(16), K(17), D(18), I(19), L(20), A(21), E(22), G(23), D(24), K(25), V(26), I(27), I(28), M(29), G(30), H(31), K(32), R(33), P(34), D(35), L(36), D(37), A(38), I(39), G(40), 
A(41), A(42), I(43), G(44), V(45), S(46), R(47), F(48), A(49), M(50), M(51), N(52), N(53), L(54), E(55), A(56), Y(57), I(58), V(59), L(60), N(61), E(62), T(63), D(64), I(65), D(66), P(67), T(68), L(69), R(70), R(71), V(72), M(73), N(74), E(75), I(76), D(77), K(78), K(79), P(80), 
E(81), L(82), R(83), E(84), R(85), F(86), I(87), T(88), S(89), D(90), D(91), A(92), W(93), D(94), M(95), M(96), T(97), S(98), K(99), T(100), T(101), V(102), V(103), I(104), V(105), D(106), T(107), H(108), K(109), P(110), E(111), L(112), V(113), L(114), D(115), E(116), N(117), V(118), L(119), N(120), 
K(121), A(122), N(123), R(124), K(125), V(126), V(127), I(128), D(129), H(130), H(131), R(132), R(133), G(134), E(135), S(136), F(137), I(138), S(139), N(140), P(141), L(142), L(143), I(144), Y(145), M(146), E(147), P(148), Y(149), A(150), S(151), S(152), T(153), A(154), E(155), L(156), V(157), T(158), E(159), L(160), 
L(161), E(162), Y(163), Q(164), P(165), T(166), E(167), Q(168), R(169), L(170), T(171), R(172), L(173), E(174), S(175), T(176), V(177), M(178), Y(179), A(180), G(181), I(182), I(183), V(184), D(185), T(186), R(187), N(188), F(189), T(190), L(191), R(192), T(193), G(194), S(195), R(196), T(197), F(198), D(199), A(200), 
A(201), S(202), Y(203), L(204), R(205), A(206), H(207), G(208), A(209), D(210), T(211), I(212), L(213), T(214), Q(215), H(216), F(217), L(218), K(219), D(220), D(221), V(222), D(223), T(224), Y(225), I(226), N(227), R(228), S(229), E(230), L(231), I(232), R(233), T(234), V(235), K(236), V(237), E(238), D(239), N(240), 
G(241), I(242), A(243), I(244), A(245), H(246), G(247), S(248), D(249), D(250), K(251), I(252), Y(253), H(254), P(255), V(256), T(257), V(258), A(259), Q(260), A(261), A(262), D(263), E(264), L(265), L(266), S(267), L(268), E(269), G(270), I(271), E(272), A(273), S(274), Y(275), V(276), V(277), A(278), R(279), R(280), 
E(281), D(282), N(283), L(284), I(285), G(286), I(287), S(288), A(289), R(290), S(291), L(292), G(293), S(294), V(295), N(296), V(297), Q(298), L(299), T(300), M(301), E(302), A(303), L(304), G(305), G(306), G(307), G(308), H(309), L(310), T(311), N(312), A(313), A(314), T(315), Q(316), L(317), K(318), G(319), V(320), 
T(321), V(322), E(323), E(324), A(325), I(326), A(327), Q(328), L(329), Q(330), Q(331), A(332), I(333), T(334), E(335), Q(336), L(337), S(338), R(339), S(340), E(341), D(342), A(343).

So, Pro residues are at positions 34, 66, 79, 110, 141, 148, 165, 255.

Now, for each Pro, I should check the preceding residue because cis peptide bonds are between the carbonyl carbon of one residue and the amino nitrogen of the next. So, the Pro would be the residue after the peptide bond in question.

Wait, actually, the peptide bond is between the carboxyl of residue n and the amino of residue n+1. So, if residue n+1 is Pro, then the peptide bond between n and n+1 could be cis.

So, I need to look for Pro at position n+1, and then check if the preceding residue (n) is such that a cis bond is favored.

But wait, the presence of Pro at position n+1 doesn't necessarily mean the bond is cis. It's just that Pro has a higher tendency to be in cis conformation when the preceding residue allows it. But in most cases, even with Pro, the bond is trans unless there's a specific reason for it to be cis, like in tight turns or specific structural motifs.

However, without structural data, it's hard to say definitively, but the presence of Pro residues increases the likelihood.

So, in the given sequence, there are multiple Pro residues. For each Pro at position n+1, the preceding residue is:

- Pro at 34: the preceding residue is R (Arginine) at 33.

- Pro at 66: preceding residue is D (Aspartic acid) at 65.

- Pro at 79: preceding residue is K (Lysine) at 78.

- Pro at 110: preceding residue is K (Lysine) at 109.

- Pro at 141: preceding residue is N (Asparagine) at 140.

- Pro at 148: preceding residue is Y (Tyrosine) at 147.

- Pro at 165: preceding residue is Q (Glutamine) at 164.

- Pro at 255: preceding residue is Y (Tyrosine) at 254.

Now, I need to consider if any of these preceding residues are Gly or Pro, but in this case, none of them are. However, Pro can sometimes form cis bonds even when the preceding residue is not Gly or Pro, especially in certain contexts like turns or loops.

But without the 3D structure, it's impossible to know for sure. However, the presence of multiple Pro residues suggests that there might be cis peptide bonds, but it's not a certainty. In the absence of specific structural information, we can't definitively say whether any of these Pro residues are in cis configuration.

Therefore, the answer would be that the protein may have cis-peptide linkages, especially around Pro residues, but without structural data, it's not certain.
</think>

The protein may have cis-peptide linkages, particularly around proline residues, but without structural data, this cannot be definitively confirmed.
\end{lstlisting}

\vspace{2.5mm}

\textbf{III. Question 2: How many hydrogen atom coordinates records does this protein have?}

\vspace{2.5mm}
\begin{lstlisting}[style=agent2, breaklines=true]
<think>
Alright, let me try to figure this out. The question is asking how many hydrogen atom coordinates records a given protein has. The protein's primary sequence is provided, which is a long string of amino acids. 

First, I know that each amino acid in a protein is made up of several atoms. Typically, an amino acid has a certain number of hydrogen atoms. For example, glycine, the simplest amino acid, has two hydrogen atoms in its side chain, while others might have more depending on their structure.

But wait, when we talk about the coordinates of hydrogen atoms in a protein structure, it's a bit different. In X-ray crystallography, which is commonly used to determine protein structures, hydrogen atoms aren't usually visible because they don't scatter X-rays strongly. So, their positions are often not recorded in the PDB (Protein Data Bank) files unless under certain conditions, like high-resolution structures or if they're part of a specific interaction.

Looking at the sequence provided, it's a long sequence, but without knowing the specific structure or the resolution of the data, it's hard to say exactly how many hydrogen coordinates there are. However, generally, most PDB entries don't include hydrogen atoms because they're not resolved. If they are included, it's usually because the structure was determined at a very high resolution, and the hydrogens were either located or modeled in.

So, considering standard practice, the answer is likely that the protein doesn't have any hydrogen atom coordinates recorded because they're not typically included in the PDB entries. Unless specified otherwise, most structures don't have hydrogen positions listed.
</think>

The protein does not have any hydrogen atom coordinates recorded. Hydrogen atoms are typically not included in standard PDB files due to their low visibility in X-ray crystallography. 

Answer: 0

\end{lstlisting}

\end{document}